\documentclass[final]{cvpr}

\usepackage{etex} %
\usepackage[accsupp]{axessibility}  %
\usepackage{times}
\usepackage{epsfig}
\usepackage{graphicx}
\usepackage{amsmath}
\usepackage{amssymb}
\usepackage{fixmath} %
\usepackage{dsfont} %
\usepackage{textpos} %

\makeatletter
\@namedef{ver@everyshi.sty}{}
\makeatother

\graphicspath{{images/}}
\usepackage{alex}
\usepackage{booktabs} %
\usepackage[caption=false]{subfig} %
\usepackage{algorithm} %
\usepackage{algpseudocode} %
\usepackage{rotating}
\usepackage{enumitem} %
\usepackage{cite} %
\usepackage[font={small}]{caption}
\usepackage{soul} %

\usepackage{pgfplots}
\pgfplotsset{compat=newest}
\usepackage{tango}
\usepackage{tikz}
\usetikzlibrary{pgfplots.statistics}

\usepackage{array}
\newcolumntype{L}{>{\centering\arraybackslash}m{1.5cm}} %

\newtheorem{defn}{Definition}

\usepgfplotslibrary{colorbrewer}
\definecolor{Dark2-8-1}{RGB}{27,158,119}
\definecolor{Dark2-8-A}{RGB}{27,158,119}
\definecolor{Dark2-8-2}{RGB}{217,95,2}
\definecolor{Dark2-8-B}{RGB}{217,95,2}
\definecolor{Dark2-8-3}{RGB}{117,112,179}
\definecolor{Dark2-8-C}{RGB}{117,112,179}
\definecolor{Dark2-8-4}{RGB}{231,41,138}
\definecolor{Dark2-8-D}{RGB}{231,41,138}
\definecolor{Dark2-8-5}{RGB}{102,166,30}
\definecolor{Dark2-8-E}{RGB}{102,166,30}
\definecolor{Dark2-8-6}{RGB}{230,171,2}
\definecolor{Dark2-8-F}{RGB}{230,171,2}
\definecolor{Dark2-8-7}{RGB}{166,118,29}
\definecolor{Dark2-8-G}{RGB}{166,118,29}
\definecolor{Dark2-8-8}{RGB}{102,102,102}
\definecolor{Dark2-8-H}{RGB}{102,102,102}

\definecolor{hous}{HTML}{b88b4d}
\definecolor{green}{HTML}{79c561}
\definecolor{farming}{HTML}{ded94c}
\definecolor{trans}{HTML}{b4b4a9}
\definecolor{services}{HTML}{ff362e}
\definecolor{other}{HTML}{dbd4d3}
\definecolor{industry}{HTML}{db79c0}
\definecolor{water}{HTML}{7982db}
\definecolor{techinfra}{HTML}{303355}

\pgfplotsset{compat = 1.3,
         legend style={font=\scriptsize},
         legend cell align={left},
         legend style={cells={align=left}, draw=black!20},
         grid=both,
         grid style={dotted},
         tick style={draw=none},
         enlarge x limits=false,
         enlarge y limits=false,
         axis line style={draw=black!100},}
\pgfplotsset{ every non boxed x axis/.append style={x axis line style=-},
    every non boxed y axis/.append style={y axis line style=-}}

\newlist{inlinelist-roman}{enumerate*}{1}
\setlist*[inlinelist-roman,1]{%
  label=(\roman*),
}
\newlist{inlinelist-alph}{enumerate*}{1}
\setlist*[inlinelist-alph,1]{%
  label=\alph*),
}

\newcommand{\tablestyle}[2]{\setlength{\tabcolsep}{#1}\renewcommand{\arraystretch}{#2}\centering\footnotesize}
\makeatletter\renewcommand\paragraph{\@startsection{paragraph}{4}{\z@}
  {.5em \@plus1ex \@minus.2ex}{-.5em}{\normalfont\normalsize\bfseries}}\makeatother

\pgfplotsset{compat=1.11,
    /pgfplots/ybar legend/.style={
    /pgfplots/legend image code/.code={%
       \draw[##1,/tikz/.cd,yshift=-0.25em]
        (0cm,0cm) rectangle (3pt,0.8em);},
   },
}

\usepackage[pagebackref=true,breaklinks=true,letterpaper=true,colorlinks=true, citecolor=blue,bookmarks=false]{hyperref}

\makeatletter\renewcommand\paragraph{\@startsection{paragraph}{4}{\z@}
 {.5em \@plus1ex \@minus.2ex}{-.5em}{\normalfont\normalsize\bfseries}}\makeatother

\def\cvprPaperID{7004} %
\def\confYear{ICCV 2021}
\newcommand{\cc}[1]{\textcolor{black}{#1}} %

\newcommand{\KG}[1]{\textcolor{black}{#1}} %
\newcommand{\KGnew}[1]{\textcolor{black}{#1}} %
\newcommand{\KGcr}[1]{\textcolor{black}{#1}} %

\newcommand{\KH}[1]{\textcolor{black}{#1}} %
\newcommand{\KHnew}[1]{\textcolor{black}{#1}} %
\newcommand{\KHthree}[1]{\textcolor{black}{#1}} %

\newcommand{\KGiccv}[1]{\textcolor{black}{#1}} %

\renewcommand{\paragraph}[1]{\vspace{2mm}\noindent\textbf{#1}}

\begin{document}

\title{From Culture to Clothing:\\Discovering the World Events Behind A Century of Fashion Images}

\author{Wei-Lin Hsiao\\
UT-Austin\\
{\tt\small kimhsiao@cs.utexas.edu}
\and
Kristen Grauman\\
UT-Austin\\
{\tt\small grauman@cs.utexas.edu}
}

\twocolumn[{%
\renewcommand\twocolumn[1][]{#1}%
\maketitle
\thispagestyle{empty} %
\vspace*{-10mm}
\begin{center}
  \centering
  \includegraphics[width=\linewidth]{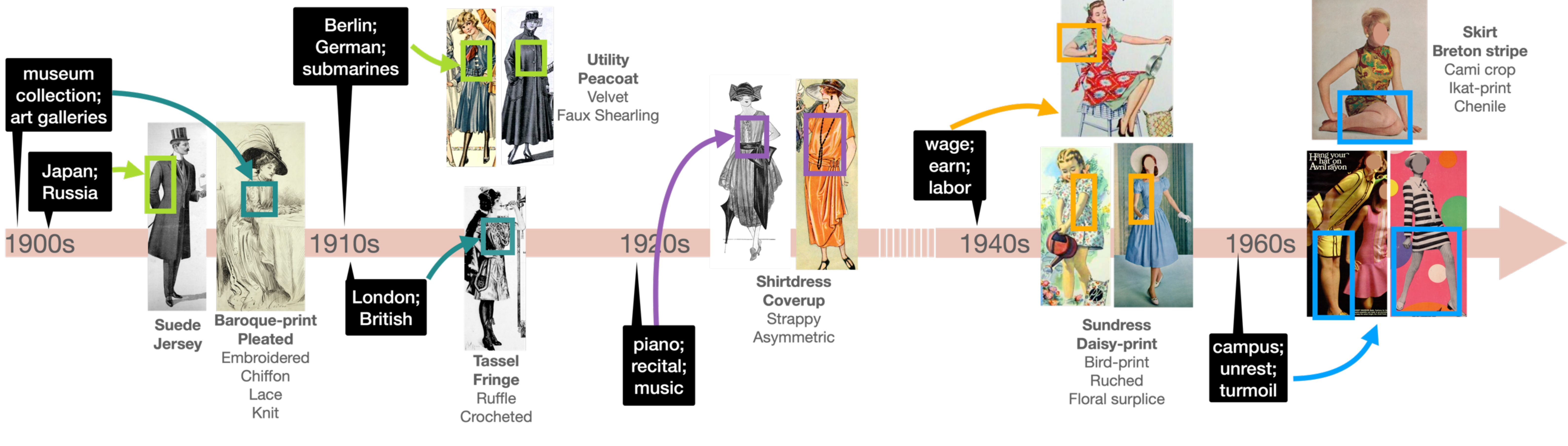}
  \vspace*{-9mm}
  \captionof{figure}{\textbf{Fashion history timeline} created by our model. We automatically discover iconic styles in each era, and detect the world events or cultural factors that gave rise to their popularity. For example, during the 1910s, World War I (\ie war-related activities in England and Germany) influenced utility clothing. On the other hand, postwar in the 1920s, people's increasing engagement in music and entertainment gave birth to the flapper style. Words listed in call-out boxes are the top words in the textual cultural topic (topic discovery described in Sec.~\ref{sec:factor_mining}). Images and the words below are the iconic styles and \KG{detected clothing} attributes/categories at that time (details in Sec.~\ref{sec:style_discovery}).}
  \label{fig:concept}
\end{center}%
}]

\begin{abstract}
Fashion is intertwined with external cultural factors, but identifying these links remains a manual process limited to only the most salient phenomena.  We propose a data-driven approach to identify specific cultural factors \KGnew{affecting}  the clothes people wear.  Using large-scale datasets of news articles and vintage photos spanning a century, 
we present a multi-modal statistical model to detect \KGnew{influence} relationships between happenings in the world and people's choice of clothing. 
Furthermore, on two image datasets we apply our model to improve the concrete vision tasks of visual style forecasting and photo timestamping. %
Our work is a first step towards a computational, scalable, and easily refreshable approach to link culture to clothing.

\end{abstract}

\section{Introduction}

Fashion is much more than Paris runways and Instagram darlings.  Fashion is what all people wear---how they express their identity, taste, interests, and even mood.  At the same time, fashion is a reflection of our society and its cultural influences.  Studying clothing trends over long periods of time offers anthropologists a treasure trove of information about how people's clothing is affected by happenings in the world, whether economic, political, or social.  

The effects are felt from both localized and broad sources.  On the one hand, one-off momentous occasions or iconic individuals can cause ripples of fashion changes, such as the publicity surrounding Titanic's voyage that attracted American designers to European fashions in the 1910s~\cite{titanic-fashion}, or the oversized sunglasses donned by Jackie Kennedy that continue to inspire women decades later~\cite{jackieo}.   On the other hand, longer periods of a collective experience can also shape trends, such as U.S. women of the 1950s forgoing boxier wartime styles for softer lines and full skirts~\cite{fashion-history}, or some African-Americans embracing ethnic Afro hairstyles following the civil rights movement~\cite{afro}, or today's shift towards comfortable casual wear in the midst of the COVID pandemic and work-from-home conditions. %

While such links are intriguing, today they
require people cognizant of both clothing  and cultural trends to manually tease them out.  This process requires expert knowledge and expensive labor, making it difficult to scale. As a result, such analysis is not updated frequently, and it focuses mostly on a few major salient styles (like those referenced above), ignoring many more subtle or localized ones.

We propose to automatically discover which events and which cultural factors influence the evolution of fashion styles (see Fig.~1).  Our idea is to associate changing style trends observed in photos with their preceding world events.  
To that end, we explore how photos and news articles can together %
tell the story of why certain clothing elements ebb and flow---or spike and disappear---over time.
Specifically, we investigate temporal %
\KGnew{relations} \KGcr{and statistical influences between} %
latent topics discovered from 100M news articles and fashion styles mined from both i) Vintage, a new dataset spanning a century of vintage photos and ii) GeoStyle~\cite{geostyle}, a massive set of social media photos spanning recent years. %
Upon discovering influential events, we demonstrate their significance by posing style forecasting and photo timestamping tasks. On two datasets, we show that modeling the culture-clothing link enables more accurate forecasts of visual styles' future popularity and more accurate dating of an unseen clothing photo.

The key novelty of this work is to formalize and automate the process of tying cultural factors to fashion.  To meet that goal, we devise  technical components on the computer vision and statistical modeling side, such as a robust visual style discovery pipeline based on body regions and a multi-modal Granger causality module. %
To our knowledge, we are the first to quantify the links from cultural trends to fashion trends in image data, and the first to demonstrate the value for forecasting and timestamping.   We show the former improves over existing  methods in the literature. 
In addition, we show examples of the discovered influences, both dominant and subtle. Beyond providing better forecasts \KG{and timestamping}, our approach is a step towards a computational, scalable, and easily refreshable way to understand how external factors affect the clothes we wear.

\vspace{-2mm}
\section{Related work}
\vspace{-2mm}

\paragraph{Clothing recognition and styles.}
Computer vision for fashion often starts with %
retrieving similar garments~\cite{street-to-shop2012,where-to-buy-iccv2015,kuang2019fashionpyramid}, 
 detecting their characteristics (color, pattern, or shape \emph{attributes}) and categories (dress, blouse)~\cite{darn,ddan,deepfashion,fashionpedia}, or segmenting individual garments%
 ~\cite{paperdoll,ATR,modanet,imp},
which are all essential for product search~\cite{whittle-ijcv,dialog-retrieval,augmented-memory}.
\KG{Beyond recognition, recent work infers how well one garment matches another~\cite{mcauley-dyadic,han-mm2017,craft-ambrish,shih2017compatibility,weilin-cvpr2018,VasilevaECCV18FashionCompatibility}, 
how fashionable an outfit is as a whole~\cite{fashionability,weilin-iccv2019},
and which garments flatter which body shapes~\cite{what-dress-fits-me,weilin-cvpr2020}.}
While most prior work learns clothing \emph{styles} with %
supervision~\cite{hipsterwars,takagi2017styledataset,deepfashion},
styles can also be mined automatically~\cite{weilin-iccv2017,ziad-iccv2017,snavely-street-style,geostyle,fcdb}. 
Our model begins by recognizing garments' attributes and categories, then automatically discovers localized visual styles. \KG{However, unlike any of the methods above,} we analyze style changes through time in the context of cultural events.

\paragraph{\KG{Visual} trend analysis \KG{and dating photos}.} 
In addition to clothing styles~\cite{weilin-iccv2017,ziad-iccv2017,snavely-street-style}, styles are also of interest in other visual phenomena like automobiles~\cite{lee2013car} and architecture~\cite{doersch2012paris}.
Earlier work uses hand-engineered features (\eg, %
HOG descriptors~\cite{hog}) to mine for localized part patches (\eg, car headlights, building windows, shoulder vs.~waist regions) that are %
\KG{visually} consistent but temporally discriminative, 
then tracks styles' transitions through time and/or space~\cite{lee2013car,doersch2012paris,runway-realway}. Several prior datasets gather photos annotated by their dates in order to track trends. 
For example, the car dataset~\cite{lee2013car} spans cars from year 1920 to 1999; 
the US yearbook dataset~\cite{ginosar2015yearbook} spans faces from 1905 to 2013; a clothing dataset~\cite{vittayakorn2017when} spans 1900 to 2009.\footnote{\cc{This dataset is not publicly available.}}
The above methods focus on tracking discovered styles' transitions and predicting the \KG{year or decade} the subject (car or person or clothing) was from. 

\KG{Our work also entails modeling visual styles over time, but with the unique goal of automatically grounding style trends in world events.  While the yearbook project~\cite{ginosar2015yearbook} discusses links to social and cultural happenings (e.g., prevalence of glasses, curvature of people's smiles) and points to literature supporting those plausible influences, the links are anecdotal and found manually.} 
\KG{In contrast}, we \emph{automatically detect} cultural factors that influence the styles, 
and we use the detected influences to improve two quantitative tasks---forecasting and timestamping.

\paragraph{Influence modeling.}
While fashions %
spread to other places and %
people, only limited prior work explores modeling fashion's \emph{influence}.  Hypothesizing a runway to real-way pattern of influence, one method \KG{anecdotally} tracks the style trend dynamics of three attributes (floral, neon, pastel)\cite{runway-realway}, while another  %
monitors the correlation coefficient\KG{---but not influence or causality---}of attribute changes for the NY fashion show and the public within the same year~\cite{devil-ny}.  
\KG{The GeoStyle project identifies anomalies in an Instagram style's trendline, then \KGnew{looks at} %
text in the corresponding image captions \KGnew{to explain them} (e.g., finding a sudden burst of yellow shirts in Thailand for the king's birthday)~\cite{geostyle}.}
\KG{Beyond correlations, we discover the influence of a root source that precedes a fashion event.}
To our knowledge, the only work in computer vision utilizing detected influences are two recent style forecasting methods\cite{ziad-cvpr2020,knowledge-forecast2020}.
They model fashion influences between cities~\cite{ziad-cvpr2020} or between multiple styles and their taxonomy~\cite{knowledge-forecast2020}, \KG{with no reference to cultural events}.  %
All the above methods only consider influence relations within visual content, whereas we study the influence of external factors \KG{(news events)} on visual styles.

\section{Approach}
We first introduce the Vintage clothing image dataset and textual corpus we collected for this problem (\secref{data_collection}). 
We then describe our model for mining clothing styles across a century (\secref{people_clothing} and \secref{style_discovery}), 
and discovering cultural factors from news articles (\secref{factor_mining}). 
Finally, we detect which cultural factors influence which clothing styles (\secref{influence_modeling}), 
and apply the discovered influences to forecast future clothing trends (\secref{influence_forecast}) \KG{and timestamp photos (\secref{timestamp})}.  

Throughout, we perform parallel experiments using our new Vintage dataset---which offers a long historical window on culture---as well as \KGcr{GeoStyle~\cite{geostyle}}, an existing large-scale dataset of social media photos---which offers a rich record of cultural effects in recent years.

\subsection{\KG{Collecting a century of data}}
\label{sec:data_collection}
\vspace{-2mm}
The twentieth century saw the most rapid evolution in clothing so far. 
Mass-production techniques were introduced, 
and people's roles in society changed. 
Clothing designs \KGcr{became %
freer in form} than in previous centuries. 
Moreover, %
photography became popular beyond professionals in the late 19th century~\cite{home-portraiture}, 
generating more records of people's daily outfits. 
\KG{For all these reasons}, %
we select the twentieth century to study clothing's evolution with society.
\cc{The methodology could be extended to other time frames, provided access to relevant data.}

\paragraph{Image data.} To construct a large collection of photos spanning the twentieth century, 
we use the online social platform Flickr, %
where users share photos of topics/subjects they love. %
\emph{Vintage} is one of the popular topics on Flickr. 
Users  upload scans of old photos, magazine covers, and posters, 
 with meta-data often describing when and by whom the  images were created. 
We use keywords related to vintage clothing to retrieve \KG{publicly available} images and their meta-data, 
and automatically parse the meta-data to obtain date labels for each image. 
\KG{The resulting Vintage image dataset contains 6,292 photos and 11,898 clothing instances in total (details below).}  It is the largest date-annotated clothing dataset publicly available,  \KG{and thanks to its community photo sharing origins, it enjoys some organic diversity:} anyone can share their photos on Flickr, as opposed to photos curated in museums or textbooks. 
\KG{That said, as with any Internet photo collection, certain biases are possible.}
\KG{In our case, sampling bias exists due to variance in} how widespread photography was at different times and locations. 
For example, the outfits are mostly Western styles, 
subjects in the images are often fashion models, movie actors, or political characters, 
and there are more images in later than earlier decades. %
\figref{concept}, \figref{feature_comparison}, and \figref{torso_style_frequency_timeline} show example images. %

In \secref{experiment}, we use %
Vintage to discover influences across $100$ years, 
and we separately use the 7M-image GeoStyle dataset~\cite{geostyle} to discover influences within 2013-2016.  The two datasets have complementary strengths for our study: the Vintage photos  cover a much longer time period and multi-media sources (personal photos, magazines, ads, etc.), whereas the GeoStyle photos densely cover several recent years with a focus on social media user photos.

\paragraph{Text data.} To obtain information on what happened, 
what people discussed most, and 
what impacted people's daily lives the most, 
an ideal source is news articles.  %
We select The New York Times to be our textual corpus. 
It contains news articles for the entire twentieth century, 
is regarded as a national ``newspaper of record'' based on authority and accuracy, 
and covers a wide range of content (\eg, whereas The Wall Street Journal is business-focused). 

We collect all available New York Times news articles from $1900$ to now, for a total of 100M articles (details below).
Being an American newspaper, its content \KG{or perspective} is often about the U.S. or Western hemisphere. 
This \KG{reasonably} matches the perspective of our collected image dataset, 
making it suitable for mining potential cultural factors that shaped clothing styles across the same time period. 
We share the collected datasets at: \url{http://vision.cs.utexas.edu/projects/CultureClothing}.

\vspace{-2mm}
\subsection{\KG{Clothing features}}
\label{sec:people_clothing}
\vspace{-2mm}
We %
first apply person detection~\cite{detectron} on all images to isolate the clothing people wore. 
This gives us in total $11,898$ clothing instances.

\KG{Next we extract clothing styles.} A style representation should capture the color, pattern, cuts, \etc, of an outfit, 
while being invariant to the pose \KG{or identity} of the underlying person or other irrelevant factors.  \KG{Furthermore,} an important trait in style evolution is that changes are often gradual and local. 
For example, as clothing's function became more practical, 
in the early 1910s it first relaxed the bust area, and then towards 1920 it raised the hemline above  the calves. 
\KG{These factors call for more than a vanilla global image encoding.}
Global features from an entire outfit are often dominated by larger regions (\eg, entire silhouette of a dress), 
which would prevent analyzing localized details (subtle patterns, sleeve type, neckline, \etc). %
Similarly, features from a neural network pre-trained on ImageNet~\cite{imagenet} could capture an object's overall texture and shape, but are insufficient for the fine-grained and localized details in clothing (\eg, necklines or hemlines).

\begin{figure}[t]
    \vspace{-3mm}
    \subfloat[ImageNet~\cite{imagenet} pre-training only (left) vs.\ DeepFashion~\cite{deepfashion} fine-tuning (right).]{
      \includegraphics[width=.2\textwidth]{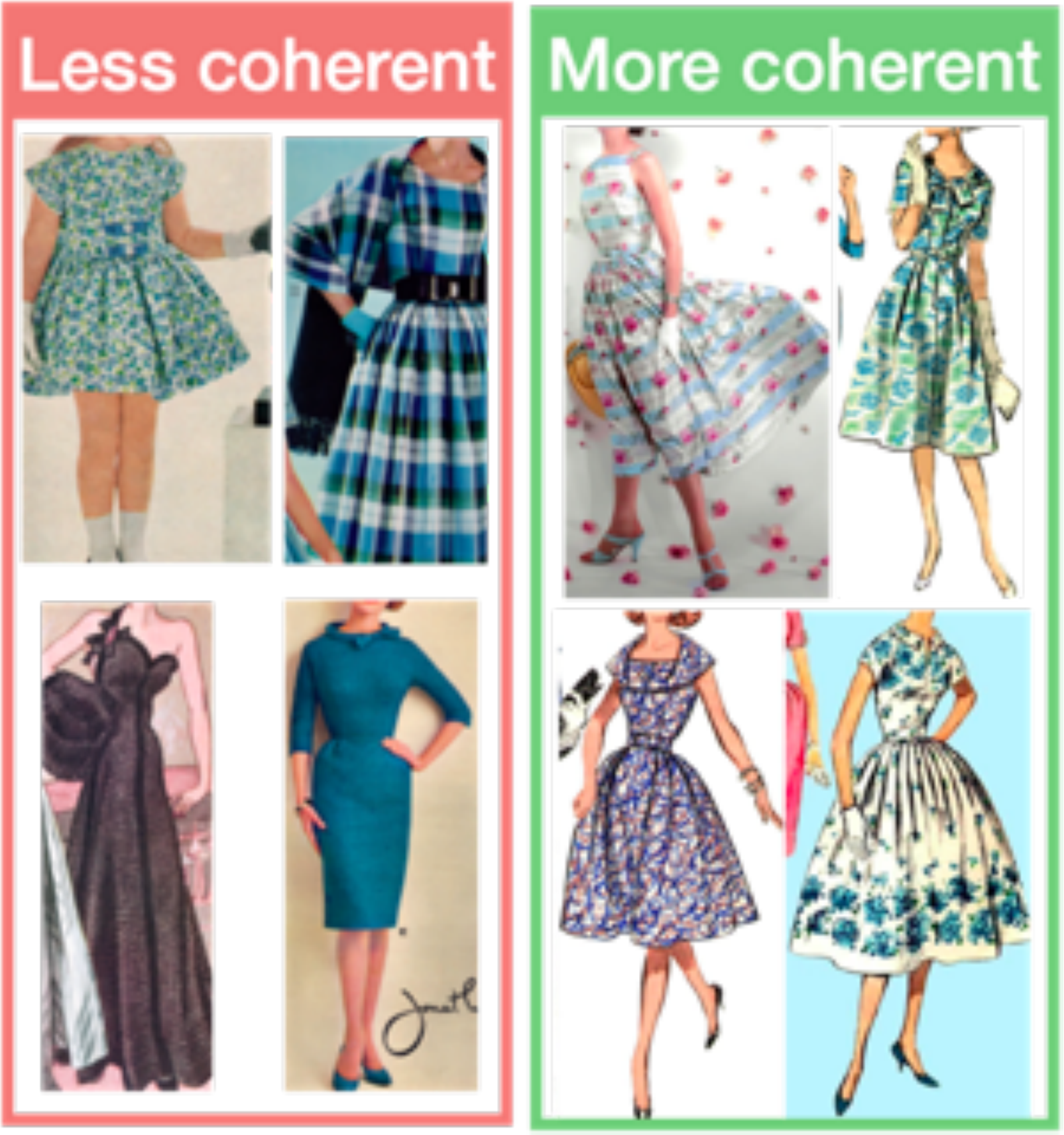}
    }\hfill
    \subfloat[Features extracted from full-body (left) vs.\ from zoomed-in body-part regions (right).]{
      \includegraphics[width=.2\textwidth]{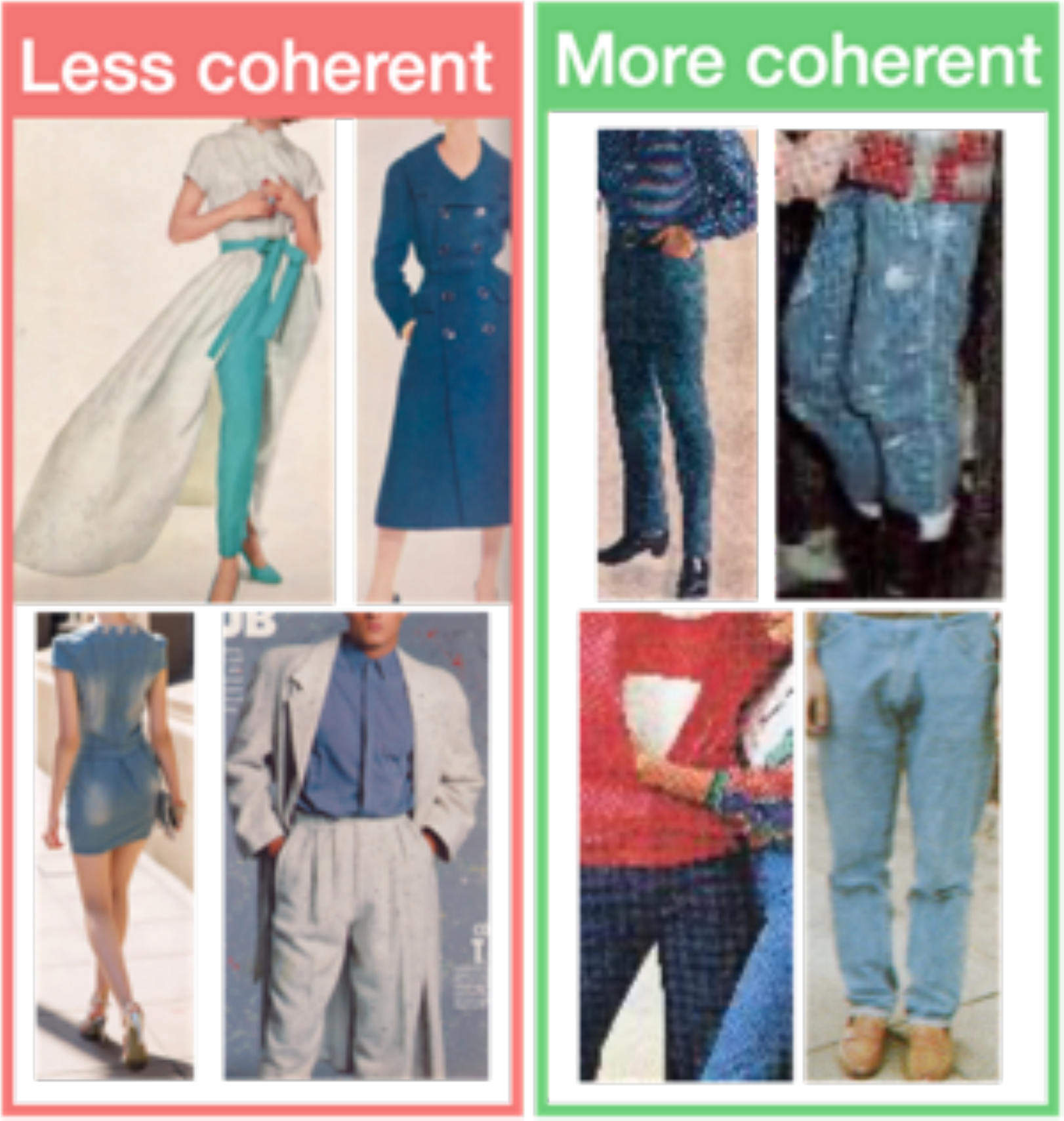}
    }
    \vspace*{-2mm}
    \caption{\textbf{Clothing features comparison}. The proposed feature extraction yields more fine-grained and coherent style clusters.}
    \label{fig:feature_comparison} 
    \vspace*{-5mm}
  \end{figure}

\begin{figure*}
    \centering
    \includegraphics[width=.99\linewidth]{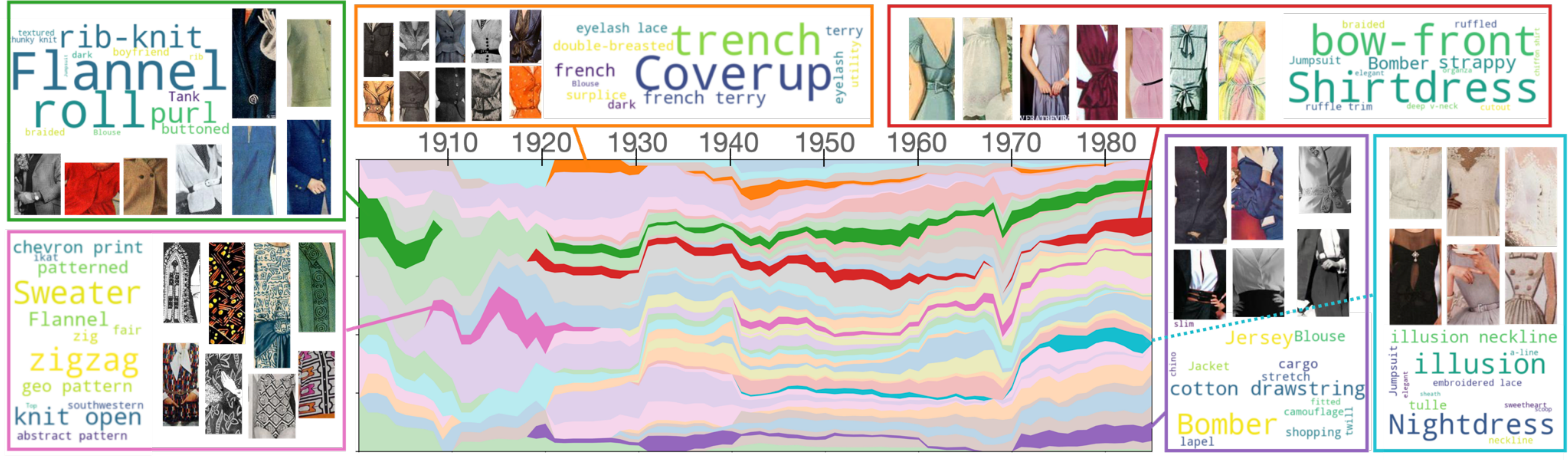}
    \vspace*{-2mm}
    \caption{\textbf{Timeline of the top styles in the torso region:} Each color represents a style, while the area a style occupies shows the frequency of that style at a time delta. Some example styles are highlighted, showing their centroid images and detected visual attributes (inferred with a classifier on the clothing-sensitive encoder). Interesting trends can be observed, e.g., styles in later times expose more skin regions than in earlier times, \KG{and some distinctive busy textures peaked in the 1910s \KGnew{and 1920s}}. See Supp.\ for timelines for other body regions.}
    \label{fig:torso_style_frequency_timeline} 
    \vspace*{-4mm}
\end{figure*}

\KG{Thus, we inject two elements to prepare the visual features to be used in style discovery.}  First, we fine-tune an ImageNet pre-trained ResNet-18~\cite{resnet} on DeepFashion~\cite{deepfashion} 
with the tasks of clothing category and attribute recognition \KG{to obtain a clothing-sensitive encoder}.
Since this network is trained to recognize details in clothing, \KG{the styles we obtain will be}
finer-grained (floral A-line dress vs.\ A-line dress). %
\KG{Second,} we zoom in on an outfit to its \emph{neckline}, \emph{sleeves}, \emph{torso}, and \emph{legs} regions, 
and analyze the evolution of clothing styles at each region separately. 
To automatically separate these regions, 
we use human body joints as anchors. 
Specifically, we detect joints for neck, arms, waist, hips, and ankles using Mask-RCNN\cite{mask-rcnn}. 
Clothing-style-based features are then extracted from these localized regions \KG{using the clothing-sensitive network.}   
\figref{feature_comparison} shows qualitative comparisons %
using the described two elements above. 
\KG{Note that the face regions in the photos are discarded; our interest is to model clothing, not identity.}  

\subsection{Clothing style discovery}
\label{sec:style_discovery}
Our goal is to mine for clothing styles
and track their trends through time \KG{as a function of world events}. 
Each clothing instance $I_j$ will thus be associated with an \KG{inferred} style label $s_j$ and a year label $d_j$. 
The year label is obtained from parsing the tag or description meta-data that comes with the image. 

To mine for clothing styles, we run clustering algorithms on the features extracted in Sec.~\ref{sec:people_clothing} for each body region. 
Since the features are already fine-tuned for clothing and localized for each region, 
using Euclidean distance as the similarity metric for clothing styles gives reasonable results. 
We use Affinity Propagation~\cite{affinitypropagation} to let the algorithm automatically decide the number of clusters for each body region. \KG{Each cluster represents a candidate style.}  %

Due to the entirely automatic process of discovering clothing styles, 
from person detection to per-part cropping based on detected body joints, \KG{as well as the presence of low-level photo statistics orthogonal to the clothing (like scanning artifacts, evolution of photo processing technologies, \etc)},
not all clusters correspond to meaningful styles. 
\KG{To control this risk,} we \KGnew{automatically} measure the quality of a cluster by how well it corresponds to \KG{common}~\cite{deepfashion} clothing attributes or categories.  %
The correspondence score for a cluster $c$ is computed as the entropy of the \KG{predicted} attribute/category label distribution:
\mbox{$E(c) = - \Sigma_{i\in S}H(i)\log_2{H(i)}$}, 
where $H(i)$ is the aggregated output activation for the attribute/category label $i$ across all instances in cluster $c$, 
and $S$ is the label set from DeepFashion~\cite{deepfashion}.
Lower entropy means better correspondence to \KG{some} %
predicted attributes/categories. 
We adopt the clusters with entropy values less than $2$ standard deviations from the mean for each body region.  \tabref{diff_only_percentage_trend_prediction} lists the final numbers of clusters (styles), \KG{which range from 26 to 144 per body region}.  

To verify the quality of the discovered styles, we conduct a user study: 
$75\%$ of the time, human judges find the clusters to exhibit coherent clothing styles that they can describe (see Supp.\ for details).

The \KG{popularity} of a style $i$ at time step $t$ is then defined as \KG{the fraction of occurrences}:
\begin{equation}
    x_{i,t} := \frac{\abr{\cbr{{s_{j'}} \mid {s_{j'}}=i, {d_{j'}}=t}}}{\Sigma_{j'}{\mathds{1}(d_{j'}=t)}},
\label{eqn:style_trend}
\end{equation}
\KG{and the trend for the $i$-th visual style over time is the sequence $x_{i,1},\ldots,x_{i,T}$, where $T$ is the most recent time point available.}
\figref{torso_style_frequency_timeline} shows the timeline of the top style trends for the torso region. See Supp.\ for the other body regions.

Note that attribute and category labels accompanying each style are for interpretability only; 
styles themselves are discovered bottom-up from clustering on clothing-oriented features, 
\emph{not} directly adopting DeepFashion~\cite{deepfashion} attributes. 
With clothing styles and their trends in hand, 
we next describe how we obtain cultural factors over time. %

\subsection{Cultural factor mining}
\label{sec:factor_mining}
To mine for the latent factors that impacted people's daily lives, 
we collect all news articles from $1900$ to %
\KGiccv{now} using the New York Times (NYT) API\footnote{\url{https://developer.nytimes.com/}},  %
then discard the extremely short ones (number of words $<15$). 
This gives us \KGiccv{$100$} million articles in total. \cc{See Supp.~for the year distribution of news articles.}
We use the concatenation of the title, abstract, and first paragraph for all articles. 
To create the vocabulary for this corpus, we use the Natural Language Toolkit~\cite{nltk} %
to apply stemming and remove stop words. 

Since most news focuses on just a few subjects, and most subjects repeatedly appear, \eg, baseball matches, presidential elections, \etc, 
we use topic modeling, specifically Latent Dirichlet Allocation (LDA)~\cite{blei2003latent}, 
to mine for the latent factors shared among all news articles.
LDA assumes that a number of latent $K$ topics account for the distribution of observed words in a document, 
where each topic is a distribution over words in the vocabulary, 
and each document can be represented as a distribution over topics. 
By running LDA (with $K=400$) over the text corpus, 
each article $M_j$ is represented as its topic distribution $\mathbold{\theta}_j$:\footnote{{\KG{To set $K$,} we use Google Cloud Natural Language API as reference, where they have hierarchical categories of common topics for news articles. There are $382$ categories in the third hierarchy.}}
\begin{equation}
    \mathbold{\theta}_j = \sbr{\theta_{j1}, \dots, \theta_{jK}},
\end{equation} 
where $\theta_{jk} \geq 0$, $\Sigma_k{\theta_{jk}} = 1$. 
Each article is dated with the date it was published, giving date label $d_j$ for article $M_j$. 
The \KG{popularity} %
of a topic $l$ at temporal bin $t$ can thus be computed as:
\begin{equation}
    y_{l,t} := \frac{\sum\limits_{j:d_j=t}{\theta_{jl}}}{\sum\limits_{k=0}^{K-1}{\sum\limits_{j:d_j=t}{\theta_{jk}}}},
\label{eqn:topic_trend}
\end{equation}
\KG{and the trend for the $l$-th cultural factor (topic) over time is the sequence $y_{l,1},\ldots,y_{l,T}$.} 

\subsection{\KGiccv{Culture-fashion} influence modeling}
\label{sec:influence_modeling}
\begin{figure}[t]
    \includegraphics[width=\linewidth]{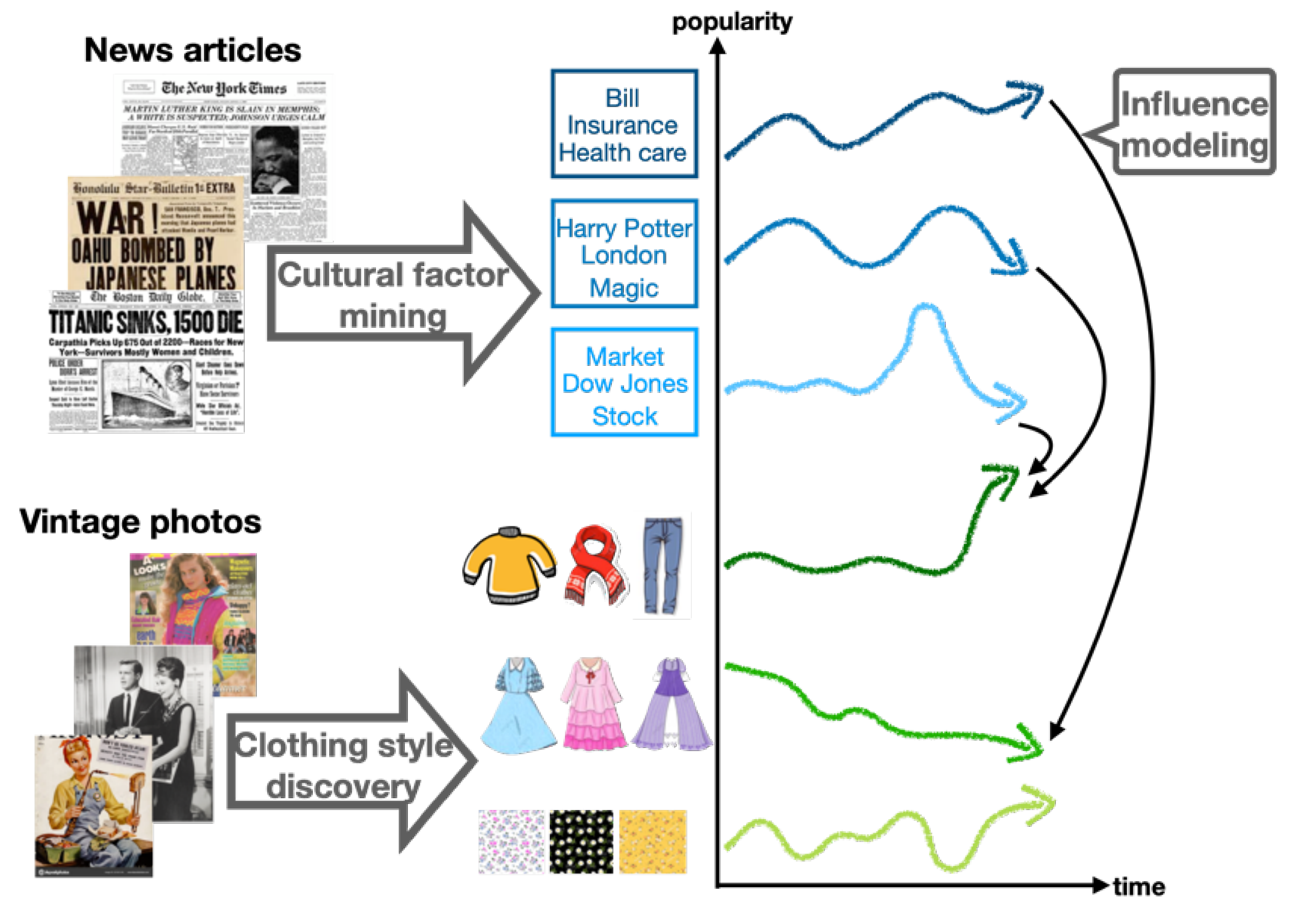}
    \vspace*{-8mm}
    \caption{\textbf{Approach overview}: Cultural factors are mined from news articles using topic models (\secref{factor_mining}), 
    and clothing styles are mined from photos by running clustering on clothing-sensitive features (\secref{style_discovery}). 
    Cultural influences on clothing styles are detected by measuring \KGnew{Granger-}causality relations between the respective popularity time series (\secref{influence_modeling}).}
    \label{fig:approach_overview} 
    \vspace*{-3mm}
\end{figure}

Having introduced our approach for discovering clothing styles and cultural factors, 
we next describe our method for \KG{modeling} how culture shapes what we wear. 
A natural thing to do would be to find the correlation between style and topic changes: 
if a style and a topic have similar changes in all adjacent years, 
it is likely that this topic influences that style.
However, this simple method fails to consider other possibilities: 
the trend of the style and the topic could be positively or negatively correlated, 
the correlation could happen at arbitrary delays, 
and local fluctuations in either trend could easily affect the overall correlation metric. 
\KGiccv{In fact, we hypothesize that} as long as \KG{observing} %
the topic helps improve forecasting the style's trend---no matter what the trends look like---that topic may have influenced the style.
This property is \KHnew{essentially} the definition of Granger causality~\cite{granger-causality}:
\begin{defn}
    Granger-causality. A time series $\cbr{y_{l,t}}$ Granger-causes another time series $\cbr{x_{i,t}}$ if including the history of $y_{l}$ improves prediction of $x_{i}$ over knowledge of the history of $x_{i}$ alone.
\end{defn}
To \KG{determine} which \KG{topic(s)} influences which \KG{style(s)}, 
we perform a Granger-causality test on all topic-style pairs, 
where the time series of topic $l$ is $\cbr{y_{l,t}}$, 
and time series of style $i$ is $\cbr{x_{i,t}}$. 
The test is based on the following regression model:
\begin{equation}
    \hat{x_{i,t}} = \sum\limits_{m=1}^{q_1}{\alpha_m x_{i,t-m}} + \sum\limits_{m=1}^{q_2}{\beta_m y_{l,t-m}}.%
\end{equation}
Here, $\alpha_m$, $\beta_m$ are the regression coefficients for each time series, 
and $q_1$, $q_2$ are their respective regression time windows.
The null hypothesis for this test is when $\beta_m=0, \forall m \in \cbr{1,\dots,q_2}$ is optimal. 
For those topic-style pairs where the null hypothesis is rejected \KGiccv{at some significance level $\alpha$,}
that topic Granger-causes the style.
\figref{approach_overview} overviews our approach.

In the following sections, 
we describe how this influence model can be used to automatically create fashion history timelines (\secref{timeline_creation}), then discuss how we verify our discovered influences through \KHthree{2} quantitative %
\KGiccv{tasks} (\secref{influence_forecast}, \ref{sec:timestamp}.)

\subsection{Automatically creating fashion timelines}
\label{sec:timeline_creation}
There are two key factors in creating a fashion history timeline: 
i) identifying iconic styles in each era, and 
ii) explaining the social and cultural happenings behind those iconic styles. 
To identify iconic styles in each decade, 
we compute the index lift~\cite{lift} for each style at each decade as $\frac{x_{i,t}}{\Sigma_{t}x_{i,t}}$, 
accounting for the uniqueness of observing that style at that time %
versus the overall style distribution across time. %
Styles with the largest lift index at time $t$ are identified as iconic styles.
To understand which happenings in the world gave rise to those styles, 
we analyze their Granger-causal topics: 
top words in a topic explain the coarser cultural factors (\eg, German, music, turmoil, \etc), 
and by tracing back the news articles that have the highest probability for that topic at that time, 
we can also %
detect the specific events (\eg, World War I, The Great Depression, campus protest and unrest, \etc) that took place.
\figref{concept} is a snippet of a fashion history timeline created with the above procedure on the Vintage data. %
While some influences shown %
may be familiar (\eg, utility clothing during wartime), 
others (mini skirts and campus unrest) are potentially newly discovered by our model.

\vspace*{-1mm}
\subsection{Influence-based style forecasting}
\label{sec:influence_forecast}
If a topic indeed influences a style, 
this %
\KGnew{relation} may continue to hold for future time series, 
thus improving the task of forecasting future style trends. 
To this end, we use our discovered influences %
in the \emph{training set} time range to help forecast style trends in the \KGiccv{future} (disjoint) \emph{test set} time range.
Trend forecasting predicts future values for a time series based on its history; autoregressive  models are typical for this task. 
Let $C_i$ be the set of influential topics for style $i$ we obtained from the Granger-causality test. 
To predict the future trend of clothing style $i$, 
we ensemble the predictions from all autoregressive models with the style's Granger-causal topics $l\in C_i$ as exogenous inputs:
\begin{equation}
    \hat{x}_{i,t} = \frac{1}{\abr{C_i}}\sum\limits_{l\in C_i}\rbrr{\sum\limits_{m=1}^{q_1}{\alpha_{i,m,l} x_{i,t-m}} + \sum\limits_{m=0}^{{q_2}-1}{\beta_{i,m,l} y_{l,t-m}}}.
\end{equation}
\KHthree{In our preliminary experiments, 
we have tried more complex models (\eg, neural-network-based), and found them to perform inferior to the simple linear-based one. 
Similar findings are reported in prior work in forecasting~\cite{ziad-iccv2017}.}

\vspace*{-1mm}
\subsection{\KG{Influence-based photo timestamping}}\label{sec:timestamp}

Aside from trend forecasting, we also examine how much the \KG{cultural factors from the} text corpus help to date (timestamp) historic photos. 
The date label of a test instance's nearest neighbor in the training set is adopted as its predicted timestamp. 
The similarity metric for the instances can be measured %
\KG{using either the visual feature alone (baseline), or by additionally using the image's \emph{inferred cultural factors} given by our model, as we describe next.}

We \KG{first train a model to map an image to its inferred cultural factors, \ie, latent textual topics.}
The textual feature $v_j$ for a training photo %
\KG{$I_j$ with date label $t$} 
is the averaged topic distribution over all news articles with that date label:
\begin{equation}
  v_j = \frac{\sum\limits_{i:d_i=t}{\mathbold{\theta}_i}}{\Sigma_i{\mathds{1}(d_i=t)}}.
\label{eqn:text_feature}
\end{equation}
At training time, a %
mapping function (a 3 layer MLP) is learned to transform a photo's visual feature to its textual feature. 
At test time, given only a photo, the textual feature of a clothing instance is obtained by feeding its visual feature to the learned mapping function, 
and the \KG{output} textual feature is used to measure similarity to the training instances. 
Both visual and textual similarities are measured by Euclidean distance in their respective feature spaces;
the resulting similarity is the average of visual and textual distances.  
In this way, \KHnew{given a photo,} we draw on the inferred cultural factors to enrich its encoding for timestamping.
\section{Experiments}
\label{sec:experiment}
We first show cultural influences on clothing styles discovered by our model.
We then evaluate the detected influences by how much they help in trend forecasting and timestamping for two \KGiccv{image} datasets.

\begin{figure*}[t]
  \vspace{-3mm}
  \null\hfill
  \subfloat[Topic `\emph{Women}' influences working attire. \label{fig:women}]{
    \includegraphics[width=.3\textwidth]{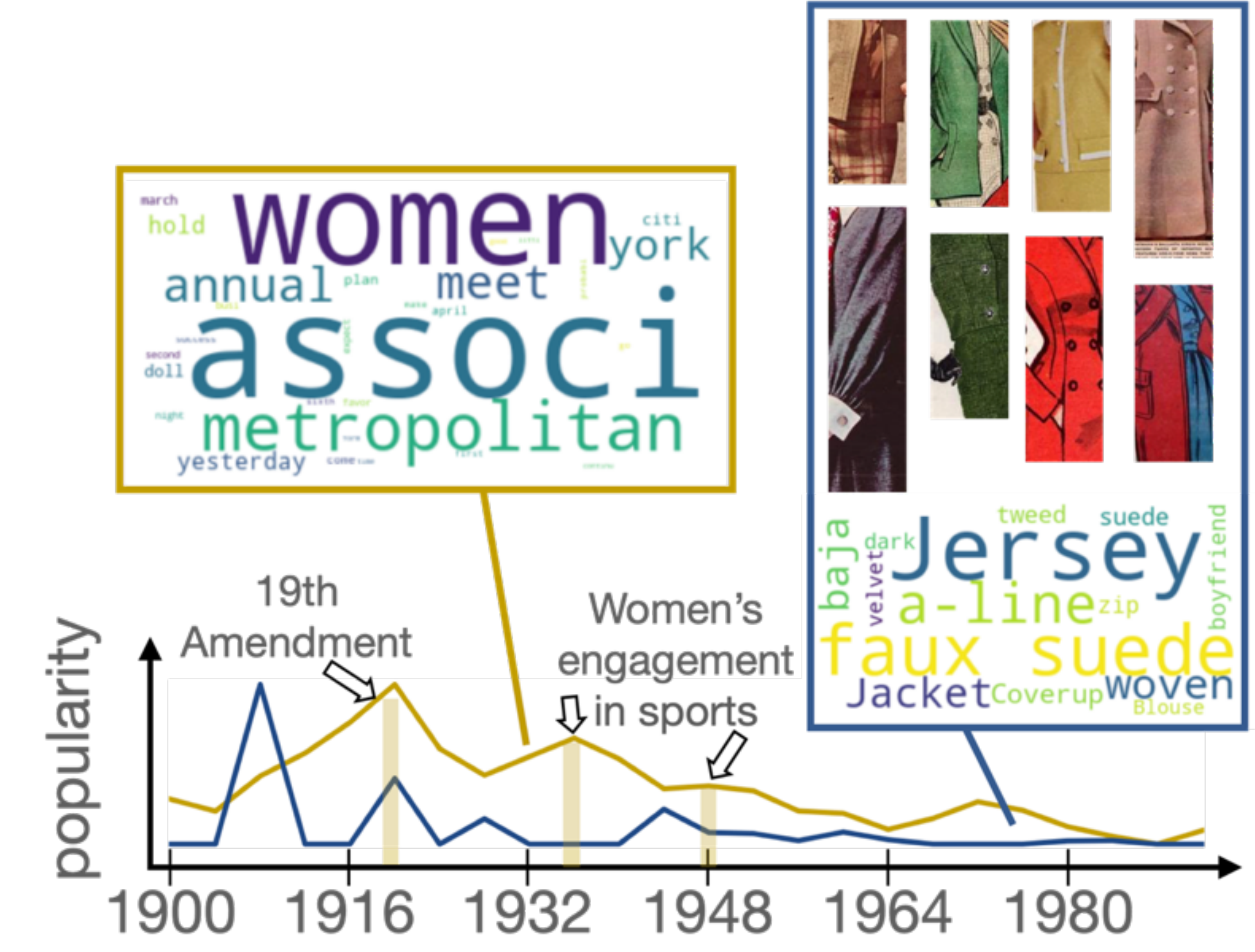}
  }\hfill
  \subfloat[Topic `\emph{War}' influences utility attire. \label{fig:war}]{
    \includegraphics[width=.3\textwidth]{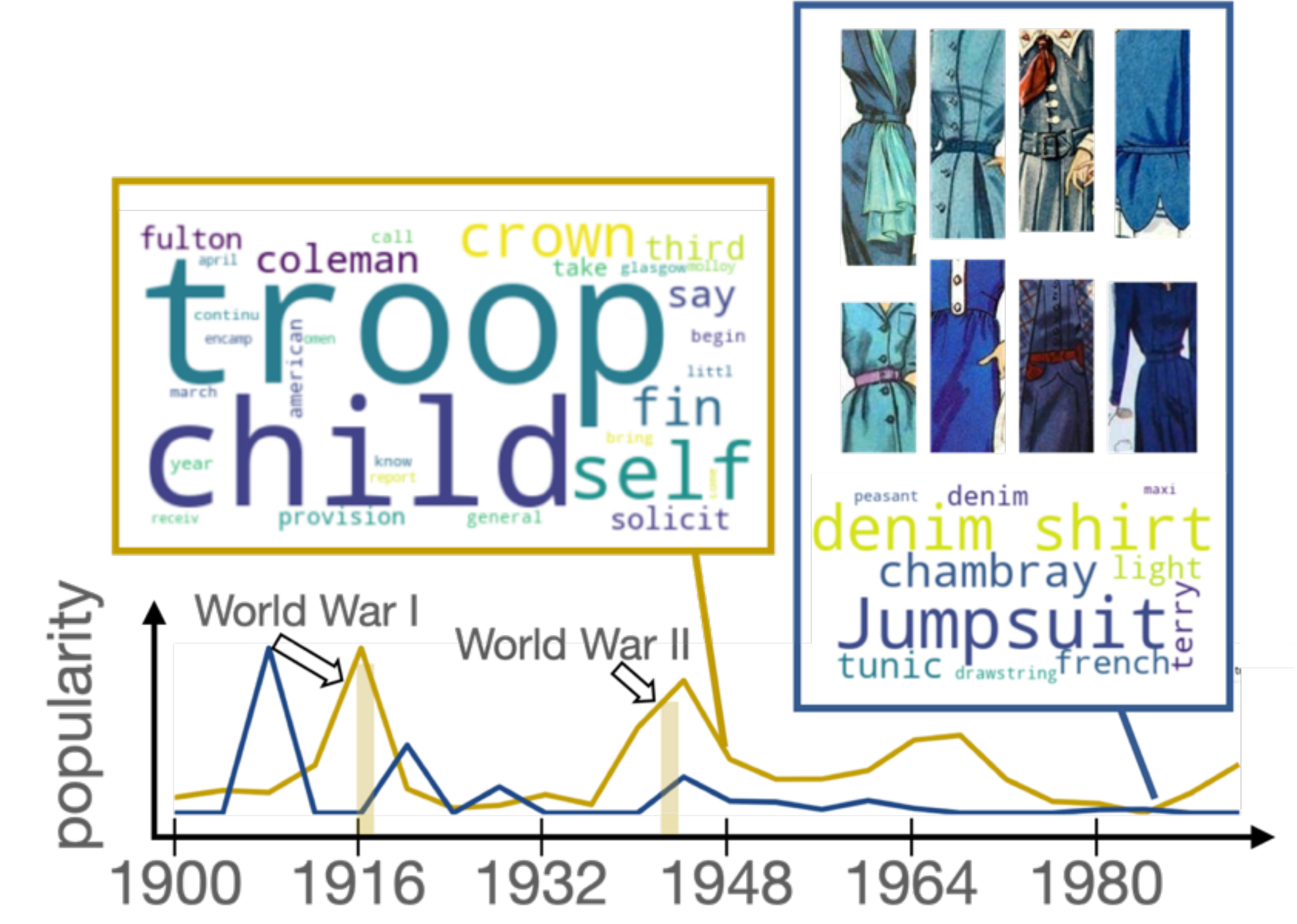}
  }\hfill
  \subfloat[Topic `\emph{African}' influences ethnic attire. \label{fig:africa}]{
    \includegraphics[width=.3\textwidth]{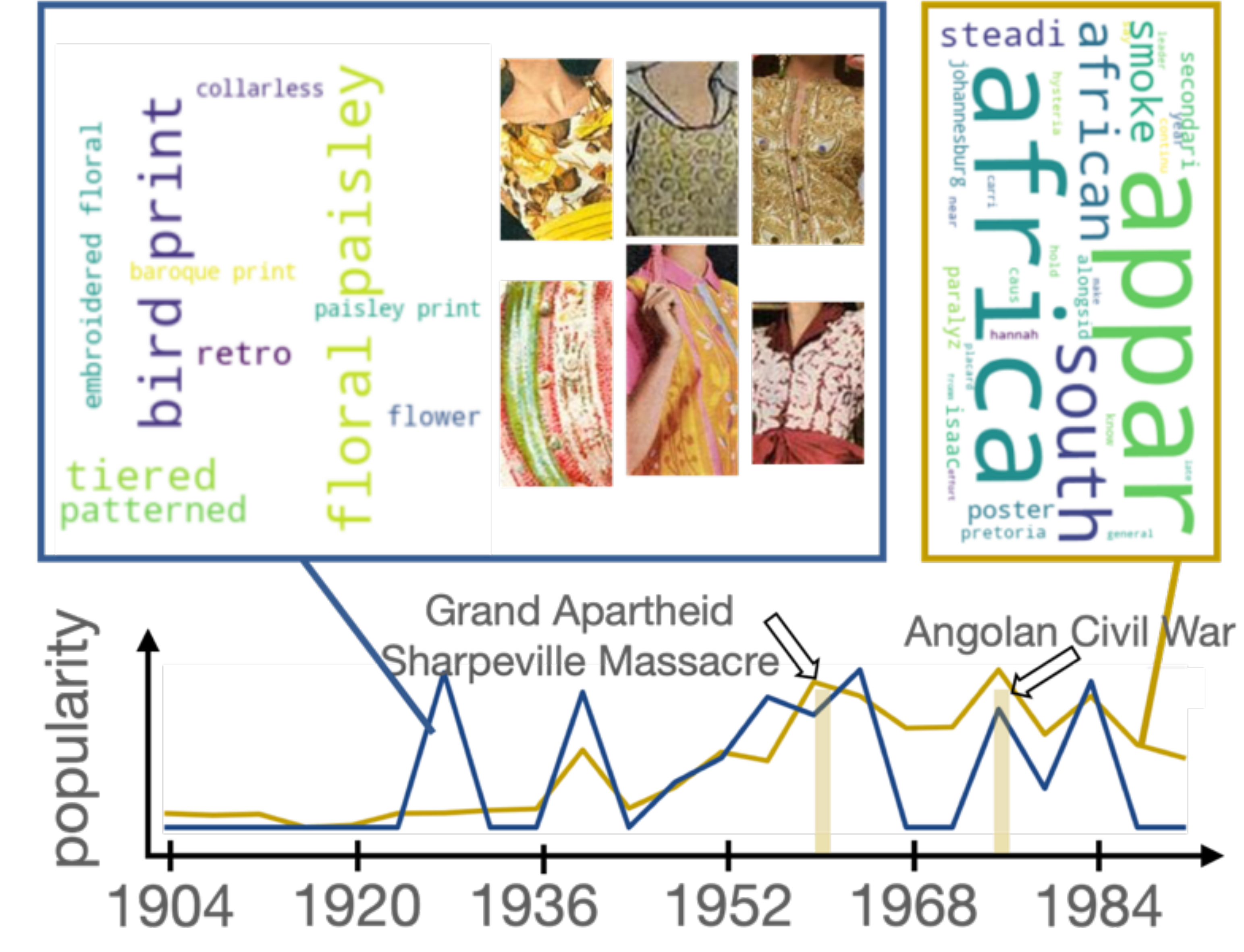}
  }\hfill\null
  \\[-2ex] %
  \null\hfill
  \subfloat[`\emph{Invention}' influences novel clothing \mbox{details}.\label{fig:patent}]{
    \includegraphics[width=.3\textwidth]{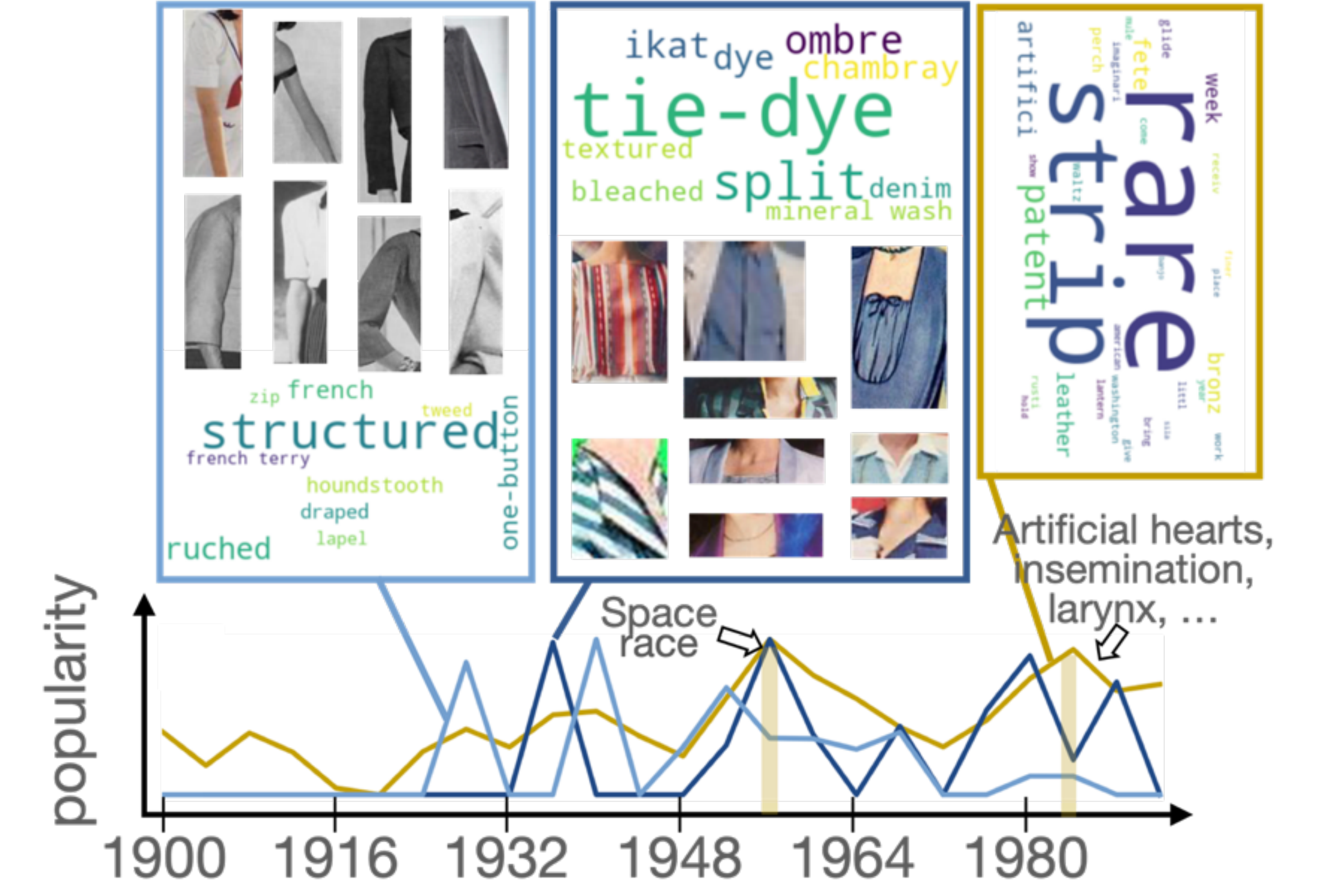}
  }\hfill
  \subfloat[`\emph{Finance}' influences necklines in \mbox{London}. \label{fig:london_folded}]{
    \includegraphics[width=.3\textwidth]{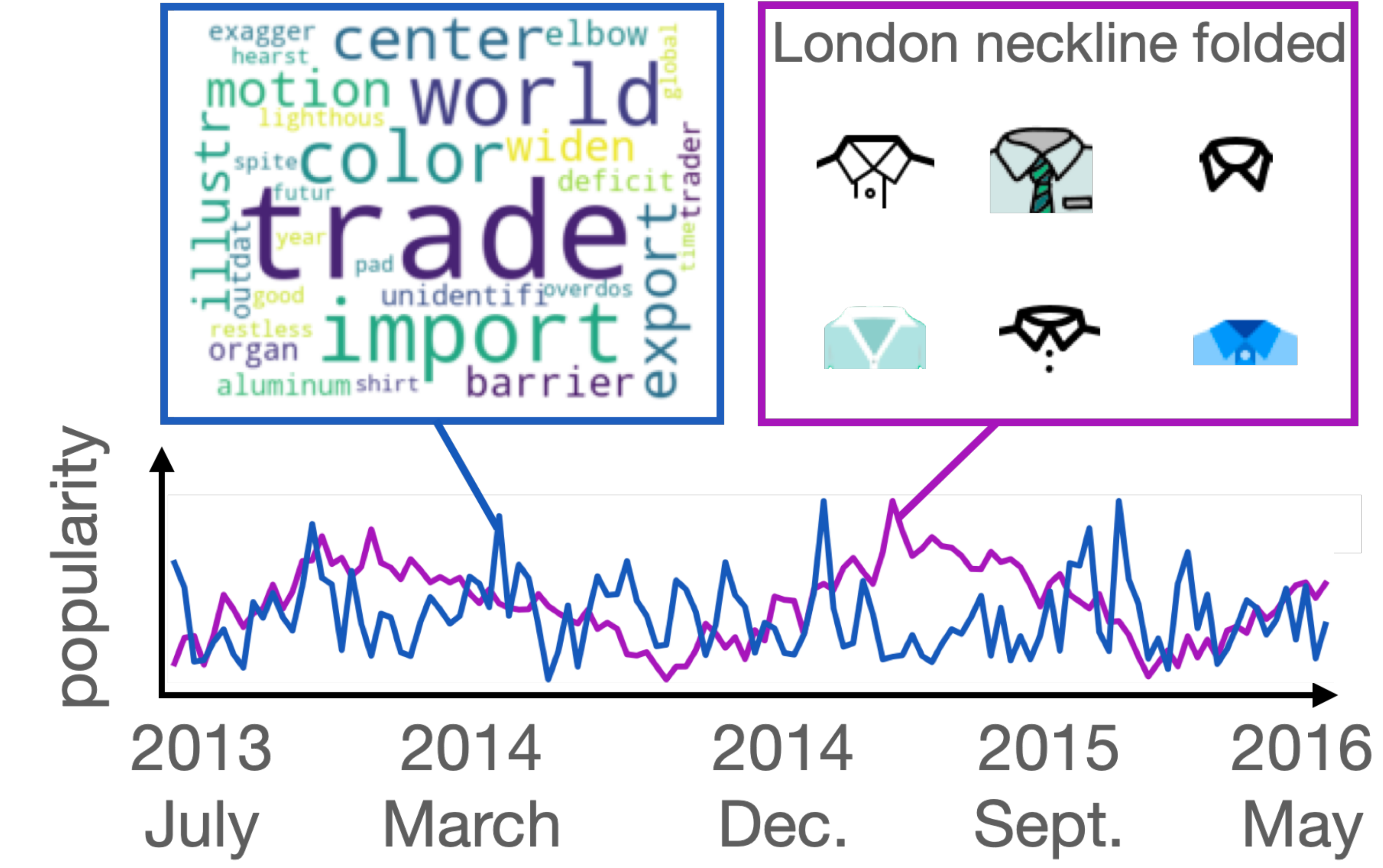}
  }\hfill
  \subfloat[`\emph{Eco}' influences wearing green in Seattle. \label{fig:green_seattle}]{
    \includegraphics[width=.3\textwidth]{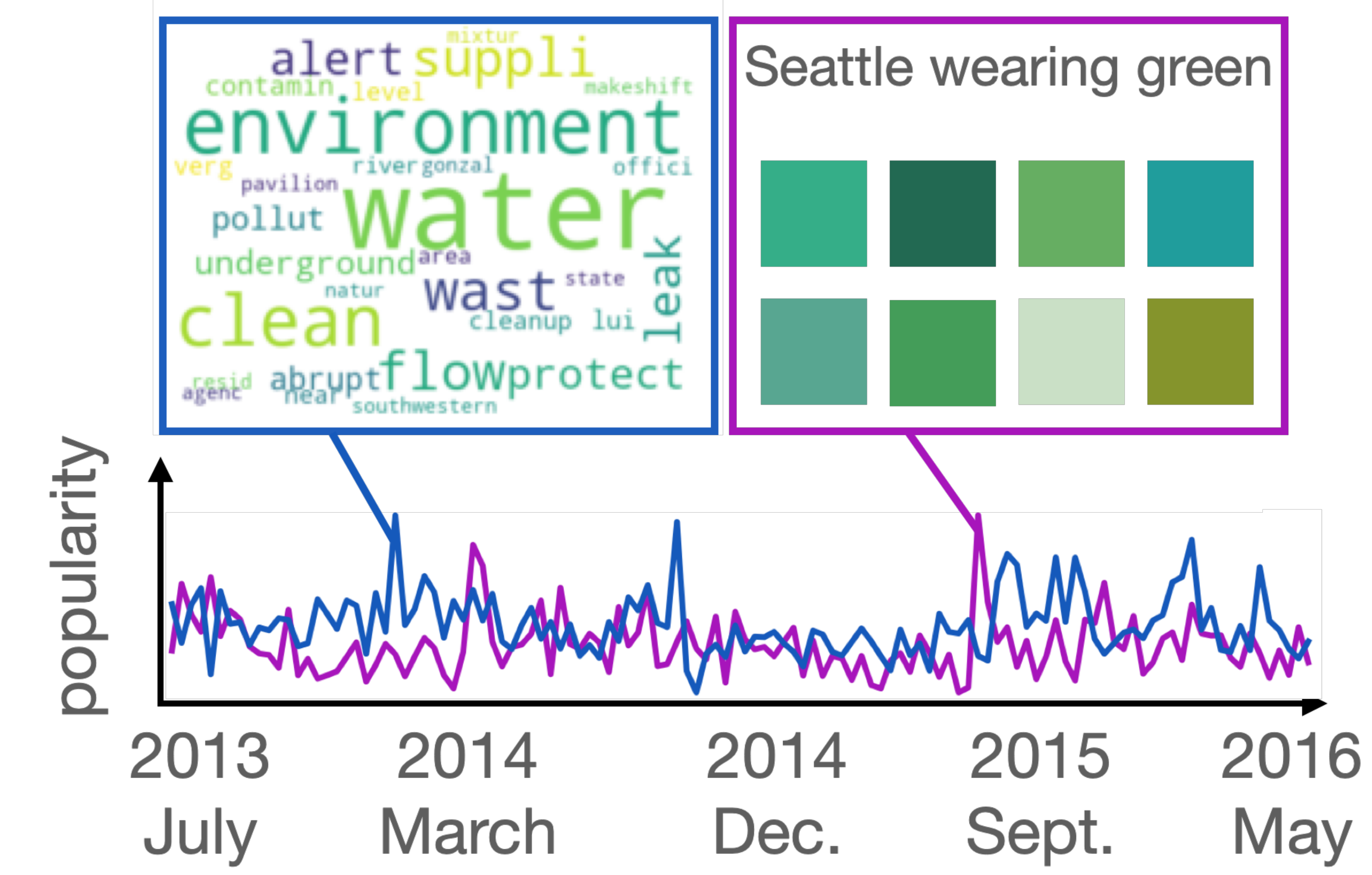}
  }\hfill\null
  \vspace*{-2mm}
  \caption{%
  \KGiccv{\textbf{Example detected influences:}} curves in each subfigure are popularity trends of visual styles and cultural factors.
           Corresponding call-out boxes for the curves show centroid images and detected attributes/categories in a style (blue boxes), 
           and also top words in a mined textual topic (yellow boxes). 
           Fig.\ (a) (b) (c) (d) show discovered  influences for the Vintage dataset (1900-1996). \cc{They agree with expert knowledge~\cite{fashion-history}.}
           Fig.\ (e) (f) show examples for GeoStyle~\cite{geostyle} (2013-2016). \KGiccv{While space permits displaying images for only a fraction of the discovered influences, our quantitative results report the results over \emph{all} discovered influences (cf.~Table~\ref{tab:diff_only_percentage_trend_prediction} and~\ref{tab:timestamp_exp}).}}
  \label{fig:discover_influences} 
  \vspace*{-3mm}
\end{figure*}

\subsection{Discovered trends and influences}

\KGiccv{First,} to study how cultural factors influence clothing styles in the long run, 
we use the Vintage data  %
(\KGiccv{cf}.~\secref{data_collection}).
The granularity of each time point $t$ in the clothing time series $\cbr{x_{i,t}}$ and textual time series $\cbr{y_{l,t}}$ 
is $4$ to $5$ \KGiccv{years}.  %
Influences are detected using %
years $1900$ to $1975$. %
The years from $1976$ %
to $1996$ are later used for evaluation on trend-forecasting. 
By using the later years as test data, where the samples are denser, 
we  assure at least hundreds of test samples per time point. 

\KGiccv{Second,} to study influence in \KGiccv{modern times} in the short term, 
we use the GeoStyle~\cite{geostyle} image data, 
which contains timestamped and geotagged photos from Instagram %
spanning a time period from July 2013 to May 2016. 
Here we use the same styles computed in previous work~\cite{snavely-street-style,geostyle}, which  aggregates detected visual attributes (at each city) weekly to obtain style trends.
This results in $2,024$ style trends, 
each of length $143$ time points. %
The granularity of each time point $t$ is $1$ week for both clothing time series $\cbr{x_{i,t}}$ and textual time series $\cbr{y_{l,t}}$.
For later evaluation on trend-forecasting, 
the last $26$ points are held out (following previous work~\cite{geostyle,ziad-cvpr2020}), 
and all previous points are used to detect influences.

\paragraph{Example influences} %
are in \figref{concept} and \figref{discover_influences}.
Cultural factors known to influence clothing styles include economic status, political tension, civil rights, wars, ethnic diversity, or new technology. 
\figref{discover_influences}(a-d) show examples of our detected influences \KG{from the Vintage photos}. 
\KHnew{Interestingly, they often agree with } %
\KG{those reported by experts}~\cite{fashion-history}. 
\figref{women} is a topic about women, which has its peak at the time the 19th Amendment was passed. 
The style influenced by this topic depicts working attire, with attributes such as blouse, suede, A-line, \etc.
The second topic in \figref{war} is about wars, which has its peaks at World War I and World War II. 
This topic influenced the popularity of utility clothing, with attributes like denim, chambray, peasant, jumpsuit, \etc.
The third topic in \figref{africa} is about (South) Africa, with its peaks at the Grand Aparthied and its civil war. 
Ethnicity-inspired clothing with colorful prints like floral or paisley along with embroidery are influenced by this topic. 
Finally, the fourth topic in \figref{patent} is about patents and new inventions, with its peak during the space-race era. 
Clothing influenced by this topic is mostly made with new techniques, 
including zipper, bleaching, mineral wash, \etc.
While it is satisfying to find influences that agree with expert insights, 
\KGiccv{our} model can also help recover more subtle and previously unexpected influences---a strength of our data-driven approach. 

\figref{discover_influences}(e-f) show examples of influences discovered by our method on GeoStyle. %
Our model discovers the seasonal similarity between a finance related topic and the folded necklines in London (\figref{london_folded}). 
It also discovers a possible causality relation from a environmentally-conscious topic to the style of wearing green in Seattle (\figref{green_seattle}).

\begin{figure*}[t]
  \begin{minipage}[c]{0.09\linewidth} 
    \includegraphics[width=\linewidth]{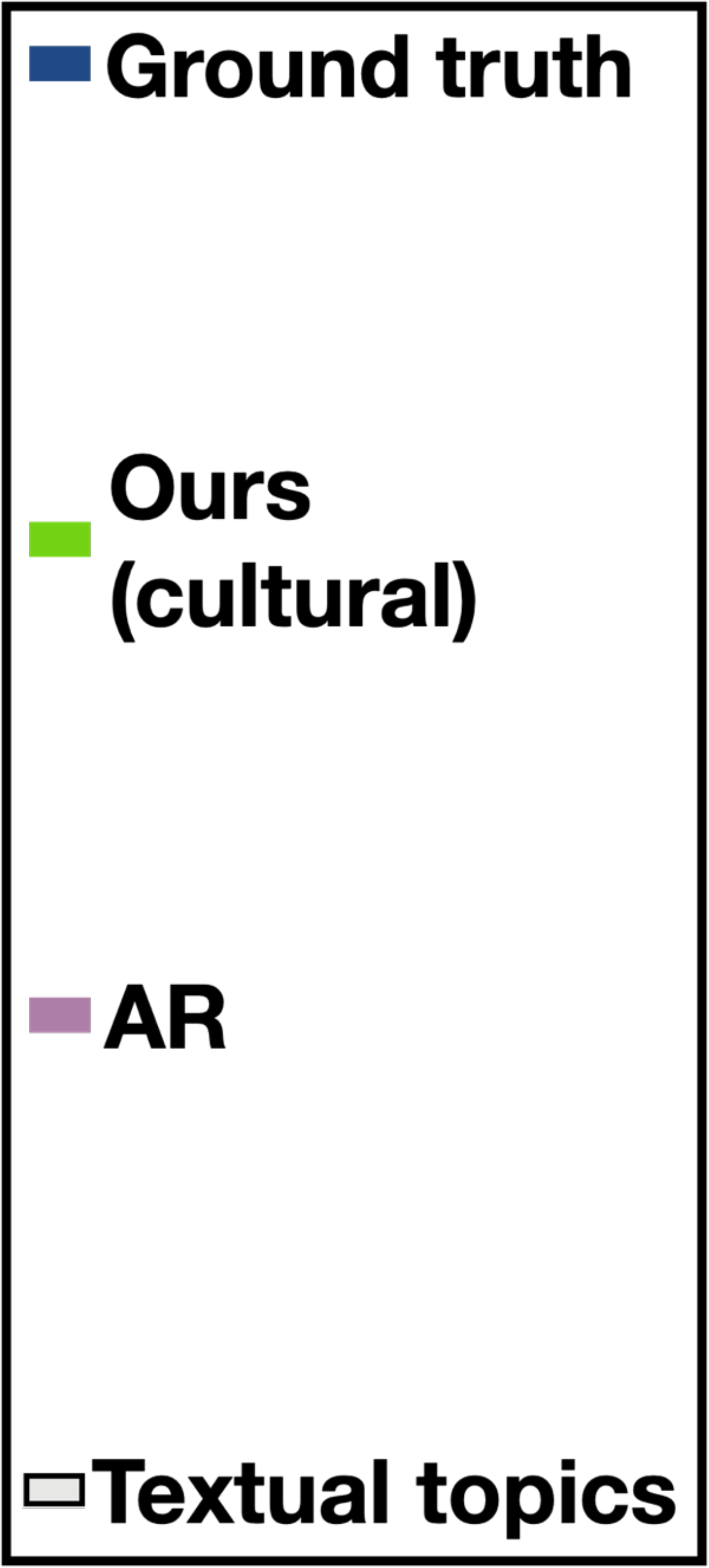}
  \end{minipage}
  \hspace{1mm}
  \begin{minipage}{0.65\linewidth}
    \includegraphics[width=.45\linewidth]{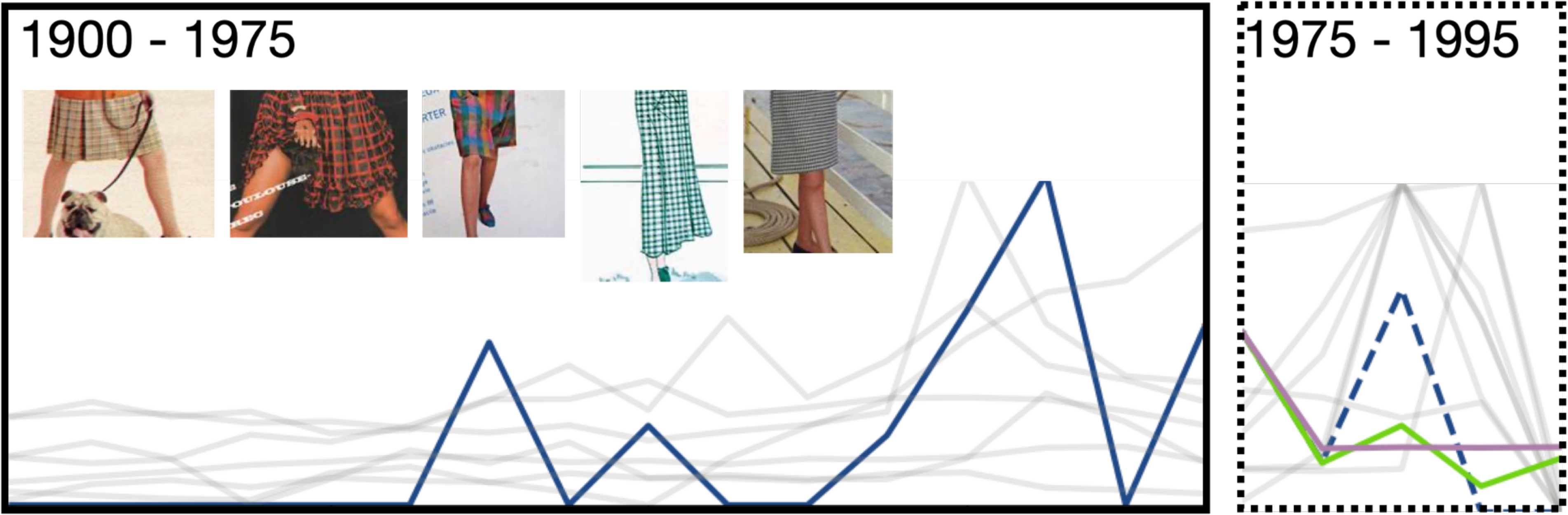}
    \includegraphics[width=.45\linewidth]{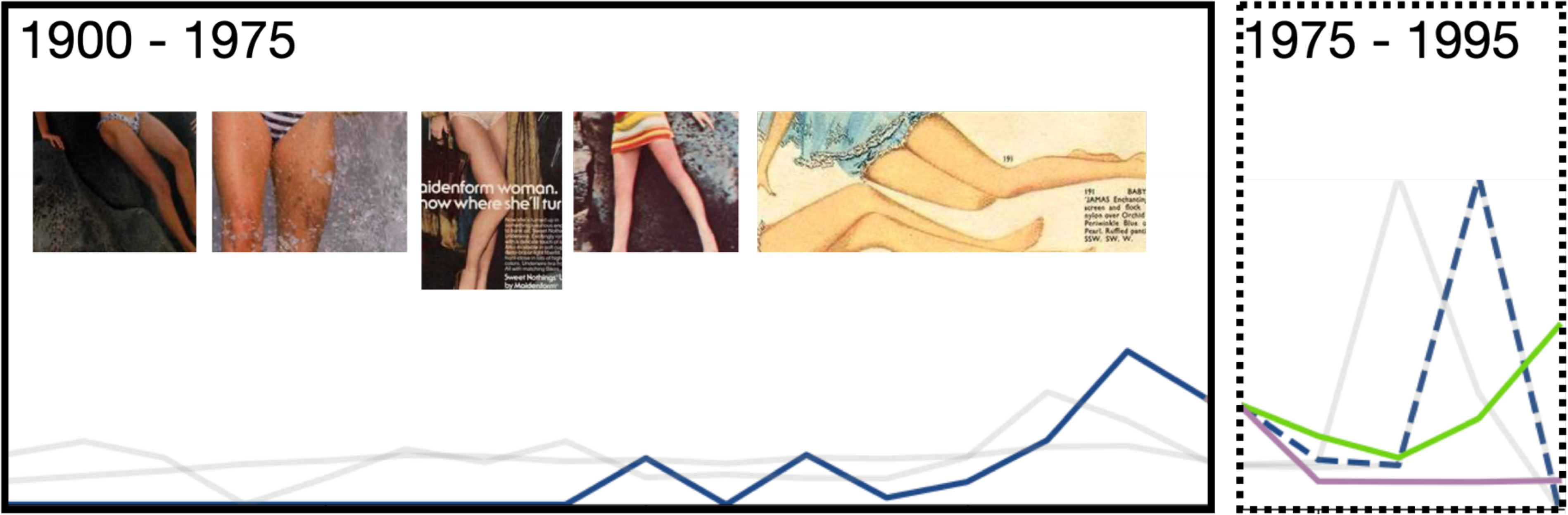}
    \\
    \includegraphics[width=.45\linewidth]{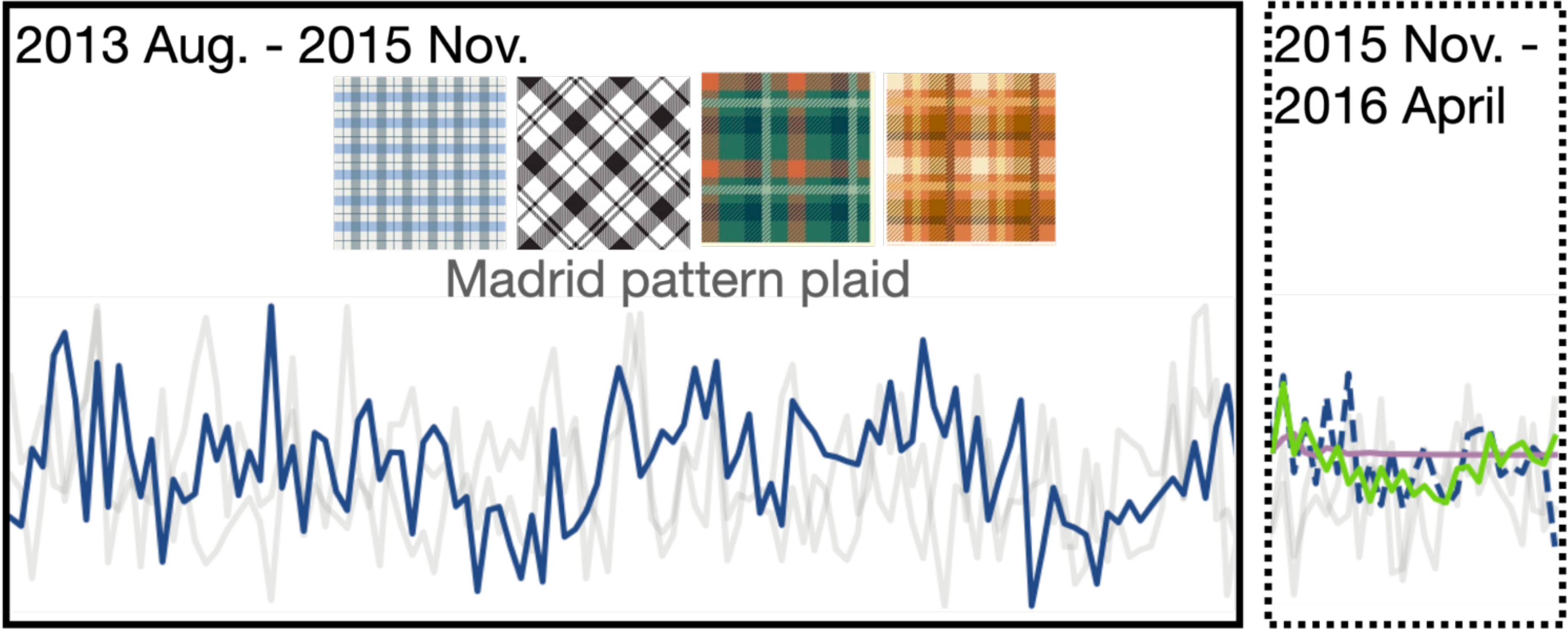}
    \includegraphics[width=.45\linewidth]{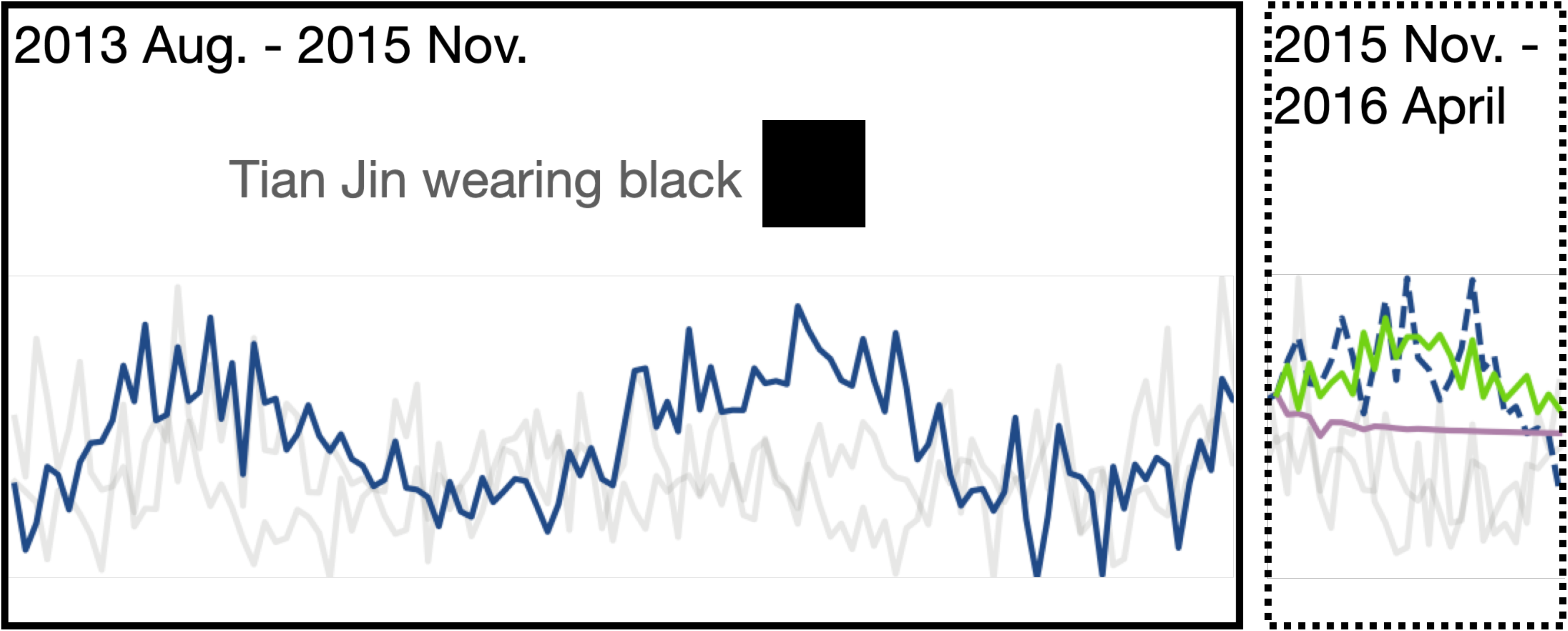}
  \end{minipage}\hspace{-10mm} %
  \begin{minipage}[c]{0.295\linewidth}  
    \caption{\textbf{Examples of trend forecasting} by vanilla AR vs.\ considering cultural influences (ours).
            \KHnew{Top row} on the Vintage data spanning a century. \KHnew{Bottom row} on the GeoStyle~\cite{geostyle} data spanning 3 years.
            The discovered \KGnew{influence} relations from topics to visual styles %
            \KHnew{help predict more accurate trends than AR}.}
    \label{fig:trend_predict_granger_ar} 
  \end{minipage}
  \vspace*{-3mm}
\end{figure*}

To verify the statistical significance of our results, 
we identify the top Granger-causality relations, 
and compare their F-values and F-critical values. F-values higher than F-critical values are considered statistically significant~\cite{ftest}. 
Vintage has an F-critical value $3.98$, and GeoStyle $2.48$. 
The top $20$ Granger-causality relations have F-values in the range $9$ to $96$ and $13$ to $17$ on each dataset, respectively, indicating they are significant.  \KGcr{See Supp.\ for full lists.}

\subsection{%
\KHnew{Forecasting trends}}
\label{sec:exp_trend_forecast}

\KHnew{
Next, we apply our detected influences on trend forecasting on the held-out time series, \KGiccv{i.e., all of 1975-1996 for Vintage, and all of the last half of 2016 %
\KGcr{for} GeoStyle.}
On both datasets, we evaluate all styles for which the external influences are adopted (\ie, those that rejected the null hypothesis in Granger tests, \KG{see Supp.}). %
The numbers of clothing styles are shown in \tabref{diff_only_percentage_trend_prediction}, first row.
}

\paragraph{Baselines and evaluation.}
We adopt the baselines from prior work on trend forecasting~\cite{ziad-iccv2017}.\footnote{%
The method of~\cite{ziad-cvpr2020} requires city labels, so is not applicable on the vintage data.  
We also tried the method of~\cite{geostyle}, 
and it fails by falling back to the linear baseline.
It could not converge when its cyclic term is included due to the lack of cyclicity in this data, which that model depends on.}
We use mean-squared-error (MSE) as evaluation metric (on vintage) because it is standard for forecasting tasks~\cite{stock2016tensor,stock2017frequency,stock2019temporal}, 
whereas we follow mean-absolute-error (MAE) for GeoStyle to be consistent with prior methods on this dataset~\cite{geostyle,ziad-cvpr2020}.
Given the ground-truth values from the training set, 
all methods predict into the future on a long-horizon basis (\ie, \KGnew{they are} never given ground-truth values from the test set).
Their performance is evaluated per style per time-point, 
and the final error is the average error over the next $20$ years and $6$ months
for Vintage and GeoStyle, respectively.

\begin{table} %
  \centering
  \tablestyle{4pt}{0.9}\begin{tabular}{@{}lc|cccc@{}}  
     & \bf{Next 6 months} & \multicolumn{4}{c}{\bf{Next 20 years}}\\
     & GeoStyle~\cite{geostyle} & neck & torso & arms & legs\\  
     styles (sig./all) & 1607/2024 & 17/74 & 86/120 & 51/144 & 2/26\\
     \toprule
     last & 0.023 & 0.303 & 0.209 & 0.258 & 0.174\\
     linear & 0.025 & 0.134 & 0.211 & 0.182 & 0.146\\
     mean & 0.034 & 0.089 & 0.085 & 0.071 & 0.149\\
     EXP~\cite{ziad-iccv2017} & 0.022 & 0.134 & 0.146 & 0.124 & 0.122\\
     AR & 0.028 & 0.091 & 0.084 & 0.070 & 0.145\\ 
     cultural (ours) & \bf{0.019} & \bf{0.081} & \bf{0.083} & \bf{0.068} & \bf{0.088}\\ 
  \end{tabular}
  \vspace*{-2mm}
  \caption{\textbf{Forecasting trends} for styles in the GeoStyle~\cite{geostyle} dataset (left) and each body region from the Vintage photos (right).}
  \label{tab:diff_only_percentage_trend_prediction}
\end{table}

\begin{table} %
  \centering
  \tablestyle{4pt}{0.9}\begin{tabular}{@{}lccc@{}}  
     & \KH{prior} &
     visual only & %
     \KG{visual + cultural (ours)}\\  
     \toprule
     Vintage & 0.170 & 0.653 & \textbf{0.695} ~(\textcolor{green}{+6.0\%}) \\ 
     GeoStyle~\cite{geostyle} & 0.121 & 0.124 & \textbf{0.156} ~(\textcolor{green}{+25.8\%})\\ 
  \end{tabular}
  \vspace*{-2mm}
  \caption{\textbf{Timestamping accuracy}:
  \KHnew{Given a photo, including its inferred cultural features helps better predict the correct date.}
  }
  \label{tab:timestamp_exp}
  \vspace{-3mm}
\end{table}

\paragraph{Trend prediction results} are presented in \tabref{diff_only_percentage_trend_prediction}.  \KG{Our method does best overall.} 
Predicting future trends in the short-term (on GeoStyle) or the long-term (on Vintage) generally requires models with different characterstics. 
When predicting future trends in the short term (next 6 months), 
explicitly depending more on recent values (\ie last and EXP) 
performs better than considering more historical values (\ie mean and linear). 
Example styles from GeoStyle like those in \figref{discover_influences} and \figref{trend_predict_granger_ar} suggest that
even though there are always local fluctuations, 
the trend of a style generally does not change drastically in a short period. 
Note how our influence-based model captures not only the overall future trend of a style, 
but also the local fluctuations %
(\figref{trend_predict_granger_ar}). 
Prediction in the long term (next 20 years) on Vintage is %
challenging.
Methods that use only $1$ or $2$ data points (\ie last or linear) to extrapolate future curves often perform poorly. 
As more historical points are considered (\ie mean, AR, EXP), the model's performance gets better.  Our influence-based model performs best by being able to predict the highly dynamic long-term future trends (as in \figref{trend_predict_granger_ar}).

In both the long- and short-term settings, 
including \KG{the proposed cultural} influences in autoregression (AR) improves the overall performance. 
On Vintage, 
\KG{$57\%$} of the styles perform better when including influences, with %
\KG{$8\%$} of the styles improving more than $10\%$.
\KHnew{On GeoStyle, $80\%$ of the styles perform better when including influences, 
with $66\%$ of the styles improving more than $10\%$.}

We stress that these experiments are comprehensive over all styles. 
While space allows showing qualitative figures for only a sample of discovered influence relationships, 
the quantitative results (\tabref{diff_only_percentage_trend_prediction}, \tabref{timestamp_exp}) are computed over \emph{all} the data. 
Furthermore, \KGiccv{we find} %
the accuracy of our discovered influences is important: 
if we incorporate all topics, 
not just the Granger-causal ones, forecasting is $30\%$ worse than vanilla AR.

\subsection{Timestamping photos}

\KHnew{Finally,} \KG{we evaluate our method's impact on timestamping unseen photos (cf.~Sec~\ref{sec:timestamp}).}
We randomly split both the Vintage and the GeoStyle\footnote{\KHnew{Full image data is not publicly available. We use $\sim130$ K images.}} 
datasets to hold out $20\%$ of the clothing instances for evaluation, and use the rest as the training database for retrieval. 
The set of \KHnew{date} labels for Vintage data is every $5$-th year, 1900, 1905, 1910, \dots, 1995, totaling $20$ labels. 
\KHnew{The set of labels for GeoStyle is every $4$-th month, from July 2013 to May 2016, totaling $10$ labels.}  
We evaluate multi-class classification accuracy.

\tabref{timestamp_exp} shows the timestamping results.  \KG{Our approach %
outperforms the visual-only baseline significantly.}  
The cultural features %
offer a better representation for timestamping than the visual features alone,
likely because they compress date-specific information in a cleaner manner.

\section{Conclusion}

\KG{Analyzing a century of fashion photos, we explored how world events may have affected the clothes people choose to wear.  Our statistical model identifies concrete temporal %
\KGnew{influence} relationships between news events and visual styles, allowing both well-known and more subtle ties to be surfaced in a data-driven manner.  To demonstrate the real-world impact, we proposed methods for both forecasting and timestamping that leverage the mined influential cultural factors.  Our results on two datasets show how this novel source of context benefits both practical tasks.  In future work, we plan to investigate hierarchical models of influence and explore the role of geographic patterns.}

\noindent \textbf{Acknowledgements:} 
We thank Greg Durrett, Ziad Al-Halah, and Chao-Yuan Wu for helpful discussions.  We also thank the authors of \cite{ziad-iccv2017} for kindly sharing their data and code with us.

{\small
\bibliographystyle{ieee_fullname}
\bibliography{strings,cvpr2019-refs,cvpr2018-refs,cvpr2020-refs,cvpr2021-refs}
}
\clearpage
{\LARGE Supplementary Material} 
\vspace{5mm}

\setcounter{section}{0}
\renewcommand\thesection{\Roman{section}}

This supplementary file contains:
\begin{itemize}
    \item Training details for all methods in trend forecasting.
    \item Style trends for legs, neck, sleeves regions.
    \item \KHnew{Top Granger-causal relations from Vintage and GeoStyle.}
    \item More examples of discovered cultural influences on clothing styles.
    \item Year distribution for Vintage dataset and New York Times dataset.
    \item Qualitative examples of timestamping through retrieval.
    \item \KH{Example failure cases in influence-based forecasting}.
    \item \KHnew{Interface and protocol for user study on quality of visual clusters.}
\end{itemize}

\begin{table*}
    \centering
    \footnotesize 
    \tablestyle{4pt}{0.9}
    \begin{tabular}{@{}l|cccccccccccccccccccc@{}} 
        Vintage & 96.6 & 36.2 & 31.7 & 24.0 & 20.6 & 17.3 & 16.8 & 14.3 & 13.9 & 13.4 & 12.3 & 11.9 & 11.6 & 11.2 & 10.2 & 12.7 & 9.9 & 9.8 & 9.4 & 9.0 \\
        GeoStyle & 16.6 & 15.8 & 15.6 & 14.1 & 14.1 & 13.9 & 13.9 & 13.6 & 13.6 & 13.5 & 13.3 & 13.3 & 13.3 & 13.3 & 13.0 & 12.8 & 12.7 & 12.7 & 12.7 & 12.6 \\
    \end{tabular}
    \vspace{-3mm}
    \caption{Top $20$ F-values for Granger-causal relations on Vintage (F-critical-value$=3.98$) and GeoStyle (F-critical-value$=2.48$). Since the F-values are greater than the F-critical value, they are statistically significant.}
    \label{tab:top20_granger}
\end{table*}

\begin{table*}
    \centering
    \footnotesize 
    \begin{tabular}{@{}cc|cc|cc|cc@{}} 
         \multicolumn{2}{c|}{Neck} & \multicolumn{2}{c|}{Torso} & \multicolumn{2}{c|}{Arms} & \multicolumn{2}{c}{Legs} \\
         Style & Topic & Style & Topic & Style & Topic & Style & Topic \\
         \toprule
         cheetah & wilson & everyday & today & southwest-pattern & wilson & skirt & clinton\\
         polka-dot & greenberg & cable-knit & earlier & tassel & greenberg & lace & southhampton \\
         dotted & jess & velvet & advertise & fringed & jess & galaxy & streak \\
         perforated & pageant & slit & settlement & geo & pageant & sheath & campus \\
         animal-print & embrace & pintuck & continue & batwing & embrace & baroque & tumultuous \\
         \midrule
         halen & crew & everyday & data & southwest-pattern & blizzard & checkered & kennedy \\
         surfer & nixon & cable-knit & change & tassel & copenhagen & grid & harold \\
         van & junior & velvet & temperature & fringed & danish & plaid & bullet \\
         love & burger & slit & wave & geo & primarily & gingham & wound \\
         palm tree & member & pintuck & phenomenon & batwing & weird & swiss & tragedy \\ 
    \end{tabular}
    \caption{Top $2$ Granger-causal relations in each body region in \textbf{Vintage}: each relation is a Granger-causality from a cultural factor (topic) to a clothing style. We show the top detected attributes/categories in each style, and the top words in each topic.}
    \label{tab:top_granger_vintage}
\end{table*}

\begin{table*}[]
    \centering
    \footnotesize 
    \tablestyle{4pt}{0.9}
    \begin{tabular}{@{}l|c|c|c|c|c|c|c|c|c|c@{}} %
         Style & Chicago-cyan & Sofia-cyan & Guangzhou-cyan & Madrid-dress & Toronto-cyan & Tokyo-dress & NYC-cyan & Chicago-cyan & Berlin-cyan & Chicago-cyan \\
         \midrule
         Topic & race & reid & wan & driver & budget & stake & office & donald & store & budget \\
               & season & seth & unmistakable & mile & year & franklin & police & rumsfeld & retail & year \\
               & finish & geography & miranda & car & cut & roosevelt & shoot & ponder & department & cut \\
               & indianna & savage & rebellion & formula & spend & crown & fatal & blumenthal & confirm & spend \\
               & kennedy & webster & dermatology & distance & billion & triplet & bronx & claudia & shop & billion \\
               
    \end{tabular}
    \caption{Top $10$ Granger-causal relations in \textbf{GeoStyle}: each relation is a Granger-causality from a cultural factor (topic) to a clothing style. Each style corresponds to a detected attribute in a city. For each topic, we show the top words in it.}
    \label{tab:top_granger_geostyle}
\end{table*}

\section{Training details for methods in trend forecasting}
In the trend forecasting experiment in Sec.\ 4.2 in the main paper, 
we compare our proposed influence-based method with $5$ other baselines. 
Here, we describe training details for each of them below. 
Let $\cbr{x_{i,t}}, t=1,\dots,T_{train}$, be the time series for style $i$ in training set. 
All methods predict future trends using the training time series. 
\paragraph{Last-baseline.} This baseline uses the immediate previous value as the predicted value for the future trend:
\begin{equation}
    \hat{x}_{i,{t+1}} = x_{i,T_{train}}, t \geq T_{train}
\end{equation}

\paragraph{Linear-baseline.} This baseline fits the training time series with a linear function (slope $m$, y-intercept $b$):
\begin{equation}
    m = \frac{x_{i,{T_{train}}} - x_{i,1}}{T_{train}-1}, b = x_{i,1},
\end{equation}
and predicts future values as:
\begin{equation}
    \hat{x}_{i,{t+1}} = mt + b, t \geq T_{train} %
\end{equation}

\paragraph{Mean-baseline.} 
This baseline aggregates the mean value from training time series, and predicts future values using the mean from training:
\begin{equation}
    \hat{x}_{i,{t+1}} = \frac{\sum\limits_{j=1}^{T_{train}}x_{i,j}}{T_{train}-1}, t \geq T_{train}
\end{equation}

\paragraph{Exponential smoothing (EXP).} 
This baseline exponentially decreases weights for past observations, 
so latest time-points have higher weights than earlier time-points:
\begin{equation}
    \hat{x}_{i,{t+1}} = \alpha x_{i,{t}} + (1-\alpha) \hat{x}_{i,t}.
\end{equation}
where $\alpha \in \sbr{0,1} $.
For our experiment, $\hat{x}_{i,t}$ is set to $x_{i,{T_{train}}}$. 
We report the best numbers for all settings by using 
$\alpha=0.30$ for the neck, torso, legs regions, 
$\alpha=0.2$ for the sleeves region on the Vintage data, 
and $\alpha=0.7$ for the GeoStyle~\cite{geostyle} data.

\paragraph{Autoregression (AR).}
Like the previous EXP baseline, 
this method also weights the past observations to predict future values. 
Instead of exponentially decreasing the weights through time, it learns weights by fitting the training series:
\begin{equation}
    \argmin_{\alpha_{i,m}, \forall m} {\normbr{\hat{x}_{i,{t+1}} - {x}_{i,{t+1}}}}^2,
\end{equation}
where
\begin{equation}
    \hat{x}_{i,{t+1}} = \sum\limits_{m=0}^{q_1-1}{\alpha_{i,m} x_{i,t-m}}, t=1,\dots,T_{train}-1,
\end{equation}
and $q_1$ is the window size that AR weights past time points in. 
$q_1$ is set to $2$ and $4$ on the Vintage and GeoStyle~\cite{geostyle} datasets. 
A window size of $2$ corresponds to $8$ ($10$) years in the vintage data, 
and a window size of $4$ corresponds to $1$ month in the GeoStyle data. 
\cc{It is safe to assume that information contained in past observations earlier than this range may already be covered in this window. 
(Past observations older than 8 or 10 years may be irrelevant for predicting yearly trends; likewise, past observations older than a month may be irrelevant for weekly trends.)}
Each style $i$ learns the regression weights $\alpha_{i,m}$s separately. 
At test time, AR makes its prediction by:
\begin{equation}
    \hat{x}_{i,{t+1}} = \sum\limits_{m=0}^{q_1-1}{\alpha_{i,m} \hat{x}_{i,t-m}}, t \geq T_{train}
\end{equation}

\paragraph{Cultural influence (ours).}
Finally, our proposed cultural-influence-based model builds on AR by including the external time series $y_{l,t}$ from mined textual topics $l \in C_i$ (described in Sec.3.6.\ in the main paper), 
where $C_i$ is the set of influential topics for style $i$. 
This set of topics is obtained by performing Granger-causality tests~\cite{granger-causality} on all style-topic pairs in the dataset. 

Like AR, this model also learns weights that best fit the training series:
\begin{equation}
    \argmin_{\alpha_{i,m}, \beta_{i,m,l}, \forall m, \forall l} {\normbr{\hat{x}_{i,{t+1}} - {x}_{i,{t+1}}}}^2, t=1,\dots,T_{train}-1
\end{equation}
where
\begin{equation}
    \hat{x}_{i,{t+1}} = \sum\limits_{m=1}^{q_1}{\alpha_{i,m,l} x_{i,t-m}} + \sum\limits_{m=0}^{{q_2}-1}{\beta_{i,m,l} y_{l,t-m}},
\end{equation}
and $q_2$ is the window size for the external time series $\cbr{y_{l,t}}$. 
$q_2$ is set to $2$ and $26$ on the Vintage and GeoStyle~\cite{geostyle} datasets, repectively. 
A window size of $26$ corresponds to $6$ months in the GeoStyle data, 
and this window size for external time series allows transferring seasonal characteristics with arbitrary lags on external series $\cbr{y_{l,t}}$ to target series $\cbr{x_{i,t}}$. 
Each style $i$ that is paired with one of its Granger-causal topics $l$ learns the regression weights $\alpha_{i,m,l}$s and $\beta_{i,m,l}$s separately. 
At test time, where $t \geq T_{train}$, this model makes prediction by ensembling predictions from all models $l \in C_i$:
\begin{equation}
  \hat{x}_{i,{t+1}} = \frac{1}{\abr{C_i}}\sum\limits_{l\in C_i}\rbrr{\sum\limits_{m=1}^{q_1}{\alpha_{i,m,l} x_{i,t-m}} + \sum\limits_{m=0}^{{q_2}-1}{\beta_{i,m,l} y_{l,t-m}}}.
\end{equation}

\section{Style trends for legs, neck, and sleeves regions}  
In Sec.3.3.\ in the main paper, 
we describe our approach for discovering clothing styles in a century of fashion images, 
and obtaining the style trends by computing their popularity trajectories. 
The timeline of the top styles in the torso region is shown in Fig.4 in the main paper. 
Here, we show the timelines for other body regions:
legs region in \figref{legs_style_frequency_timeline}, 
neck region in \figref{neck_style_frequency_timeline}, 
and sleeves region in \figref{sleeves_style_frequency_timeline}. 
Each color represents a style, while the area a style occupies shows the frequency of that style at a time delta. A general trend that seems to hold for all regions is that 
styles in later times expose more skin regions, 
like illusion necklines, off-shoulder cuts, strappy design, short skirts/pants, \etc. 
This is likely due to more liberal and open-minded views on clothing.   %

\section{Full lists of top Granger-causal relations}
To verify the statistical significance of our Granger-causal relations, 
we report the range of top F-values on both Vintage and GeoStyle data in Sec.4.2 in the main paper.
In more details, F-tests are conducted by the following procedure~\cite{ftest}: the null hypothesis is rejected if the F(-value) calculated from the data is greater than the critical value of the F-distribution for some desired false-rejection probability.  
In our setting, we use false-rejection probability $0.05$, 
which gives our model an F-critical-value $3.98$ on Vintage and $2.48$ on GeoStyle. 
The full list of top 20 F-values on both datasets is in \tabref{top20_granger}.
Since the F-values are greater than the F-critical value, they are statistically significant.
Aside from the F-values, in \tabref{top_granger_vintage} and \tabref{top_granger_geostyle} we also show the style-topic pairs of the top Granger-causal relations.

\section{More examples of detected influences}
\figref{more_discovered_influences} shows another six influences detected by our model, 
where Fig.(a-c) is on the Vintage data, 
and Fig.(d-f) is on the GeoStyle data.
\figref{finance_influence} is a topic about finance and annual reports. 
It constantly grew throughout the years, 
and influenced formal styles in skirts. 
\figref{conference_influence} is a topic about conferences, 
and had peaks when influential summits took place. 
It affects stylish-business clothing, 
like buttoned, woven sleeves, 
or herringbone-patterned collars with lapels or pin tucks. 
\figref{love_influence} is a love/romantic-centric style. 
Interestingly, it may have influenced bridal styles like sparkling and glitter designs with full, elegant skirts.

For the GeoStyle data, 
\figref{fashion_influence} is a topic about fashion (\eg, Chanel, Givenchi) and design (\eg, Karl Lagerfeld). 
It may have influenced styles of high-end dresses in Paris, France. 
\figref{music_influence} is a topic about music, and potentially influenced the popularity of wearing scarves in Milan, Italy. 
\figref{sports_influence} is a topic about sports games (\eg soccer team Chelsea). 
A number of sports teams in Europe have blue in their team colors (\eg, Barcelona, Real Madrid, Chelsea, \etc), and this topic influenced wearing blue clothing in Madrid, Spain.

\section{Year distribution on newly collected datasets}
As described in Sec.3.1.\ in the main paper, 
\KHnew{we collect new datasets for this study: image data from Flickr and news articles from New York Times (NYT). 
The year distribution of the number of instances for Vintage data is in \figref{flickr_distr}, and for NYT is in \figref{nyt_distr}. 
For both datasets, later decades have more instances than earlier ones. 
The earlier decades (prior to 1980s) are used as training, while the denser later decades (after 1980s) are heldout for testing. 
Each testing sample for the image dataset still has hundreds of clothing instances per time point.}
\begin{figure}[H]
  \centering
  \footnotesize
  \begin{tikzpicture}
    \begin{axis}[
          ybar, axis on top,
          height=4cm, width=0.47\textwidth,
          bar width=0.05cm,
          ymajorgrids, 
          x axis line style={opacity=0},
          legend style={
              at={(0.5,-0.35)},
              anchor=north,
              legend columns=-1,
              /tikz/every even column/.append style={column sep=0.05cm}
          },
          ylabel={Number of instances},
          symbolic x coords={
            1900,1901,1902,1903,1904,1905,1906,1907,1908,1909,1910,1911,1912,1913,1914,1915,1916,1917,1918,1919,1920,1921,1922,1923,1924,1925,1926,1927,1928,1929,1930,1931,1932,1933,1934,1935,1936,1937,1938,1939,1940,1941,1942,1943,1944,1945,1946,1947,1948,1949,1950,1951,1952,1953,1954,1955,1956,1957,1958,1959,1960,1961,1962,1963,1964,1965,1966,1967,1968,1969,1970,1971,1972,1973,1974,1975,1976,1977,1978,1979,1980,1981,1982,1983,1984,1985,1986,1987,1988,1989,1990,1991,1992,1993,1994,1995,1996},
        yticklabel style={
          /pgf/number format/fixed,
          /pgf/number format/precision=7
        },
        scaled y ticks=false,
      ]
      \addplot [draw=none, fill=plum3] coordinates {
        (1900,  6)
        (1901,  0)
        (1902,  0)
        (1903,  0)
        (1904,  1)
        (1905,  0)
        (1906,  0)
        (1907,  4)
        (1908,  8)
        (1909,  6)
        (1910, 20)
        (1911, 4)
        (1912, 5)
        (1913, 30)
        (1914, 8)
        (1915, 3)
        (1916, 6)
        (1917, 16)
        (1918, 4)
        (1919, 5)
        (1920, 46)
        (1921, 4)
        (1922, 12)
        (1923, 2)
        (1924, 0)
        (1925, 6)
        (1926, 4)
        (1927, 7)
        (1928, 10)
        (1929, 5)
        (1930, 189)
        (1931, 11)
        (1932, 2)
        (1933, 4)
        (1934, 14)
        (1935, 3)
        (1936, 0)
        (1937, 34)
        (1938, 108)
        (1939, 35)
        (1940, 264)
        (1941, 22)
        (1942, 7)
        (1943, 8)
        (1944, 14)
        (1945, 24)
        (1946, 11)
        (1947, 33)
        (1948, 13)
        (1949, 51)
        (1950, 696)
        (1951, 112)
        (1952, 27)
        (1953, 111)
        (1954, 225)
        (1955, 254)
        (1956, 177)
        (1957, 228)
        (1958, 257)
        (1959, 316)
        (1960, 779)
        (1961, 231)
        (1962, 122)
        (1963, 136)
        (1964, 138)
        (1965, 157)
        (1966, 276)
        (1967, 273)
        (1968, 248)
        (1969, 91)
        (1970, 116)
        (1971, 10)
        (1972, 18)
        (1973, 25)
        (1974, 23)
        (1975, 6)
        (1976, 10)
        (1977, 3)
        (1978, 136)
        (1979, 156)
        (1980, 486)
        (1981, 144)
        (1982, 495)
        (1983, 158)
        (1984, 132)
        (1985, 306)
        (1986, 398)
        (1987, 213)
        (1988, 251)
        (1989, 5)
        (1990, 0)
        (1991, 31)
        (1992, 0)
        (1993, 0)
        (1994, 0)
        (1995, 80)
        (1996, 0)};
    \end{axis}
  \end{tikzpicture}
  \vspace{-2mm}
  \caption{The distribution of clothing instances per year.}
  \label{fig:flickr_distr}
  \vspace{-3mm}
\end{figure}
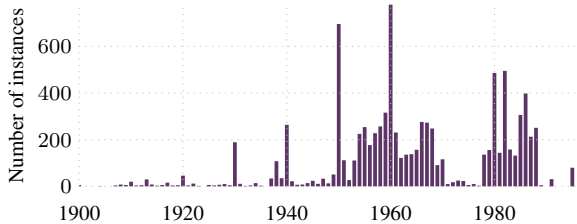

\begin{figure}[H]
  \centering
  \footnotesize
  \begin{tikzpicture}
    \begin{axis}[
      ybar, axis on top,
      height=4cm, width=0.47\textwidth,
      bar width=0.05cm,
      ymajorgrids, 
      x axis line style={opacity=0},
      legend style={
          at={(0.5,-0.35)},
          anchor=north,
          legend columns=-1,
          /tikz/every even column/.append style={column sep=0.05cm}
      },
      ylabel={Number of instances},
      symbolic x coords={
        1900,1901,1902,1903,1904,1905,1906,1907,1908,1909,1910,1911,1912,1913,1914,1915,1916,1917,1918,1919,1920,1921,1922,1923,1924,1925,1926,1927,1928,1929,1930,1931,1932,1933,1934,1935,1936,1937,1938,1939,1940,1941,1942,1943,1944,1945,1946,1947,1948,1949,1950,1951,1952,1953,1954,1955,1956,1957,1958,1959,1960,1961,1962,1963,1964,1965,1966,1967,1968,1969,1970,1971,1972,1973,1974,1975,1976,1977,1978,1979,1980,1981,1982,1983,1984,1985,1986,1987,1988,1989,1990,1991,1992,1993,1994,1995,1996},
      yticklabel style={
        /pgf/number format/fixed,
        /pgf/number format/precision=7
      },
      scaled y ticks=false,
      ]
      \addplot [draw=none, fill=plum3] coordinates {
        (1900, 30992)
        (1901, 28970)
        (1902, 29571)
        (1903, 31301)
        (1904, 27345)
        (1905, 33016)
        (1906, 35651)
        (1907, 45787)
        (1908, 44974)
        (1909, 45011)
        (1910, 43112)
        (1911, 47343)
        (1912, 52091)
        (1913, 50007)
        (1914, 52943)
        (1915, 54594)
        (1916, 51654)
        (1917, 42130)
        (1918, 44201)
        (1919, 45154)
        (1920, 52998)
        (1921, 59006)
        (1922, 72492)
        (1923, 64764)
        (1924, 67061)
        (1925, 77275)
        (1926, 72625)
        (1927, 81204)
        (1928, 77398)
        (1929, 82195)
        (1930, 81064)
        (1931, 83358)
        (1932, 80189)
        (1933, 79851)
        (1934, 86899)
        (1935, 86551)
        (1936, 91937)
        (1937, 131560) 
        (1938, 123473)
        (1939, 72350)
        (1940, 72503)
        (1941, 78807)
        (1942, 78025)
        (1943, 69709)
        (1944, 64419)
        (1945, 66537)
        (1946, 70740)
        (1947, 71155)
        (1948, 76720)
        (1949, 77023)
        (1950, 73485)
        (1951, 68788)
        (1952, 76986)
        (1953, 74170)
        (1954, 75475)
        (1955, 70914)
        (1956, 75036)
        (1957, 66068)
        (1958, 74002)
        (1959, 75380)
        (1960, 67952)
        (1961, 79536)
        (1962, 69616)
        (1963, 56363)
        (1964, 41428)
        (1965, 72108)
        (1966, 87192)
        (1967, 80573)
        (1968, 71392)
        (1969, 68178) 
        (1970, 134499) 
        (1971, 134595)
        (1972, 139887) 
        (1973, 131670) 
        (1974, 128006) 
        (1975, 126571) 
        (1976, 117232) 
        (1977, 120814)
        (1978, 89216) 
        (1979, 111889)
        (1980, 96177) 
        (1981, 135272) 
        (1982, 136239) 
        (1983, 135678) 
        (1984, 144540) 
        (1985, 139630) 
        (1986, 146140) 
        (1987, 148767)
        (1988, 146517)
        (1989, 140805)
        (1990, 135811)
        (1991, 120686)
        (1992, 118442)
        (1993, 114037)
        (1994, 113792)
        (1995, 118975)
        (1996, 112393)};
    \end{axis}
  \end{tikzpicture}
  \vspace{-2mm}
  \caption{The distribution of news articles per year. }
  \label{fig:nyt_distr}
  \vspace{-3mm}
\end{figure}
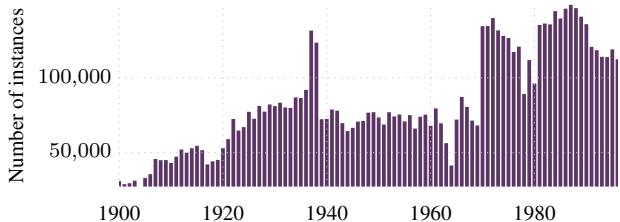

\section{Qualitative examples of timestamping through retrieval}
Quantitative results for timestamping a photo through retrieval is shown in Tab.~2 in the main paper. 
Here, we show qualitative examples comparing retrieval results using visual features only or also including inferred cultural features (approach described in Sec.~3.7 of the main paper) in \figref{timestamp_ex}. 
These are examples where visual features alone are not enough for accurately predicting query photos' date labels, 
and including cultural features helps. 
While all the retrieved photos look stylistically similar (color, pattern, fit, \etc) to the query, 
retrieved photos that include inferred cultural features are more temporally sensitive, 
with clothing styles unique to the query photo's time period (date label).

\begin{figure*}
    \centering
    \includegraphics[width=.99\linewidth]{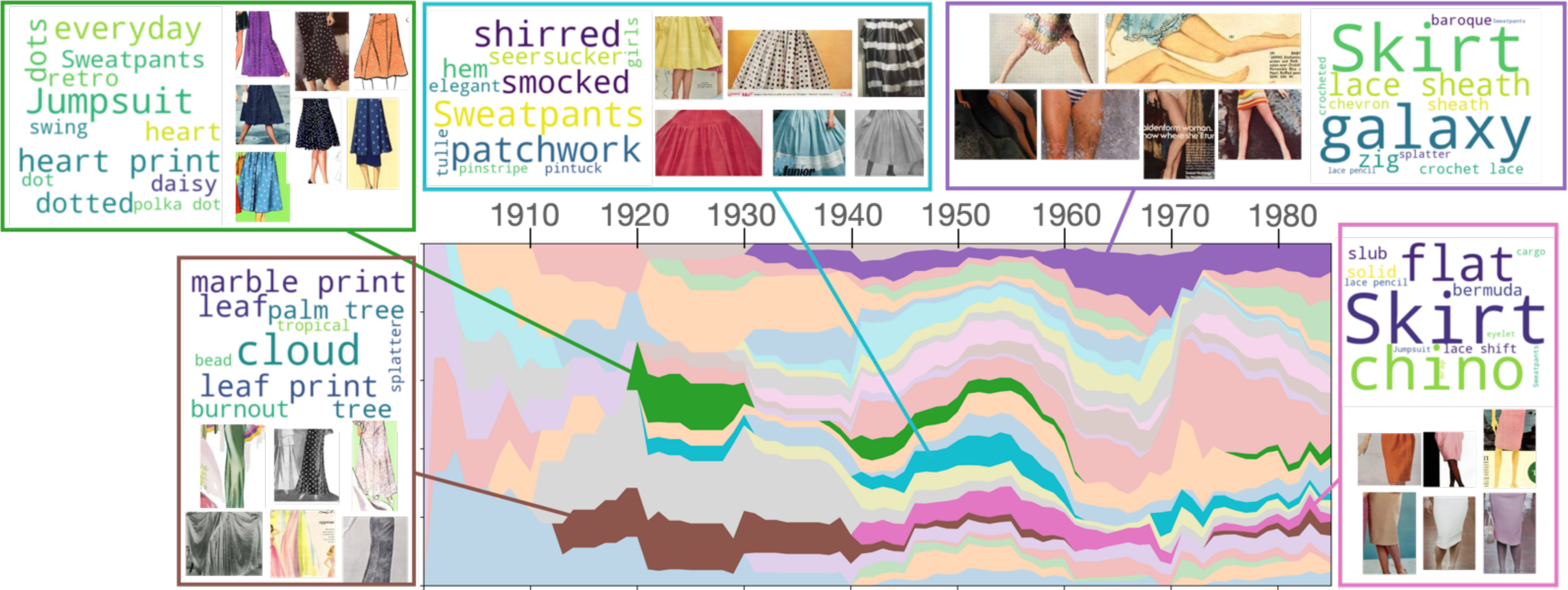}
    \caption{Timeline of the top styles in the \textbf{legs} region: Styles in later times generally accentuate more natural curves in the legs region, either exposing more skin areas (purple box) or with tighter cuts (pink box).}
    \label{fig:legs_style_frequency_timeline} 
\end{figure*}

\begin{figure*}
    \centering
    \includegraphics[width=.99\linewidth]{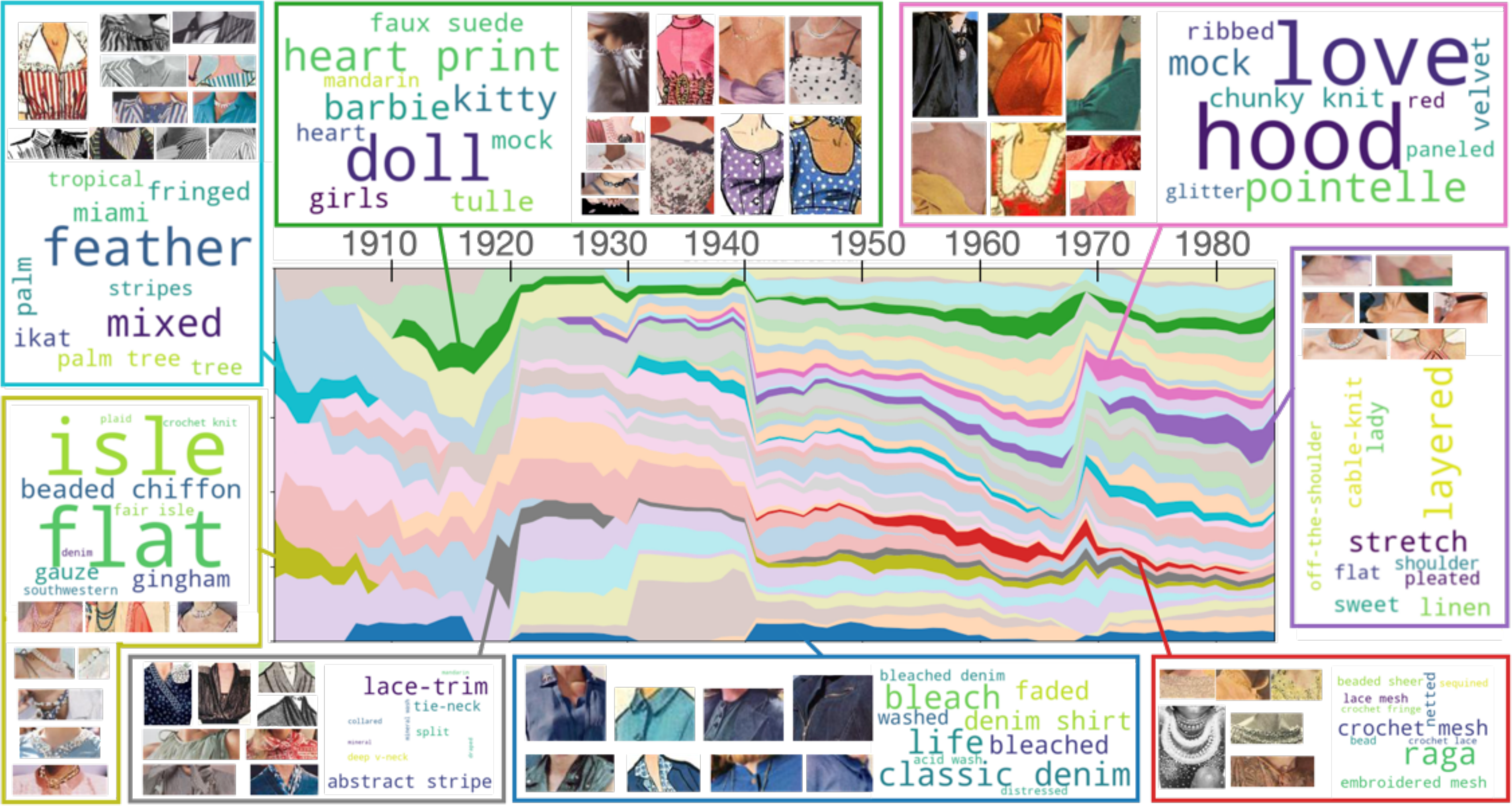}
    \caption{Timeline of the top styles in the \textbf{neck} region: Styles in earlier times generally have higher necklines (sky-blue box and olive-green box), while those in later years have deeper cuts (purple box and pink box).}
    \label{fig:neck_style_frequency_timeline} 
\end{figure*}

\begin{figure*}
    \centering
    \includegraphics[width=.99\linewidth]{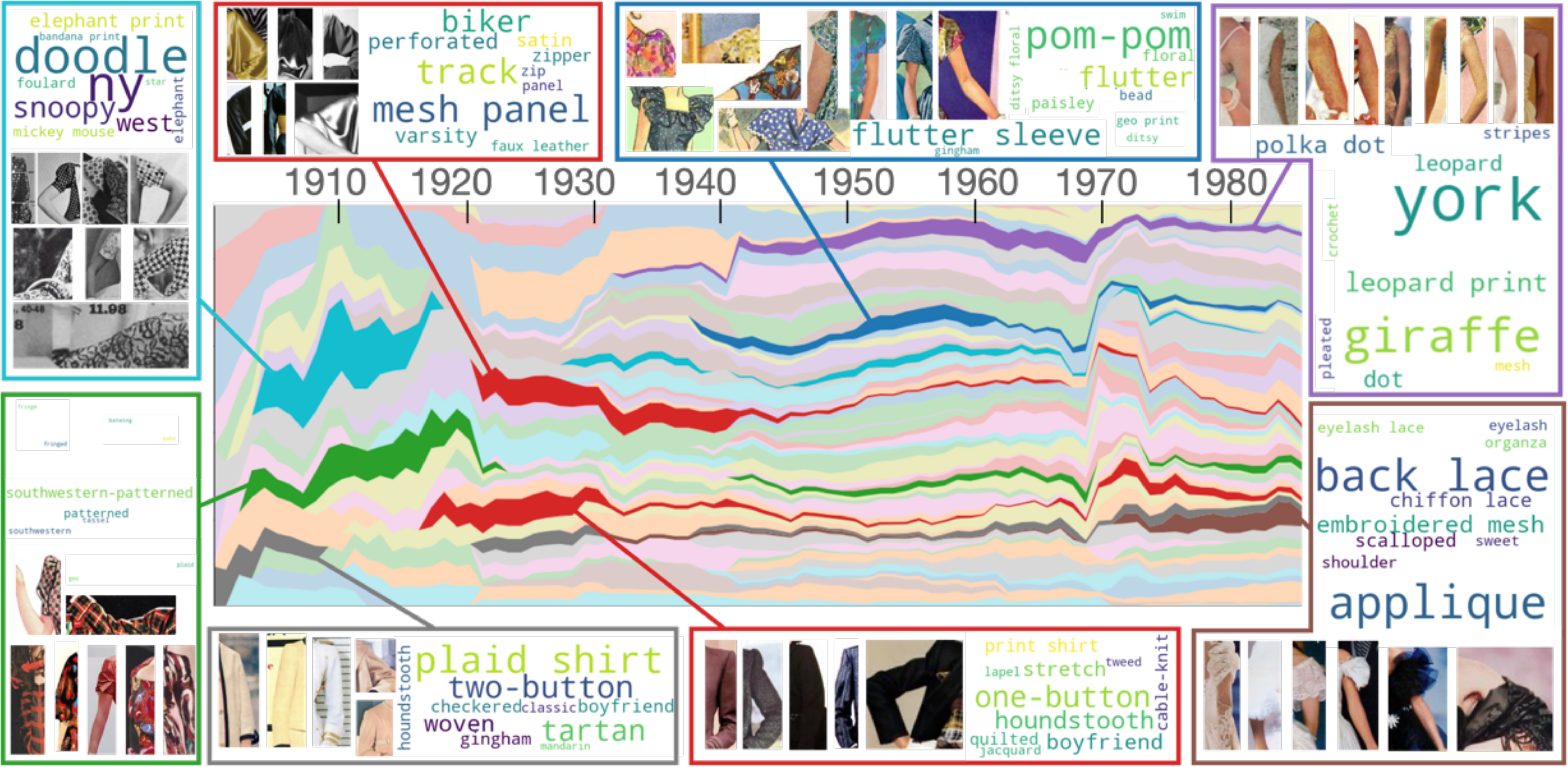}
    \caption{Timeline of the top styles in the \textbf{leftarm} region: Styles in earlier times have more busy textures but conventional cuts (green box and sky-blue box), while those in later times have fancier cuts but solid patterns (purple box).}
    \label{fig:sleeves_style_frequency_timeline} 
\end{figure*}

\begin{figure*}
  \subfloat[Topic `\emph{Finance}` influences formal skirts.\label{fig:finance_influence}]{
    \includegraphics[width=.31\linewidth]{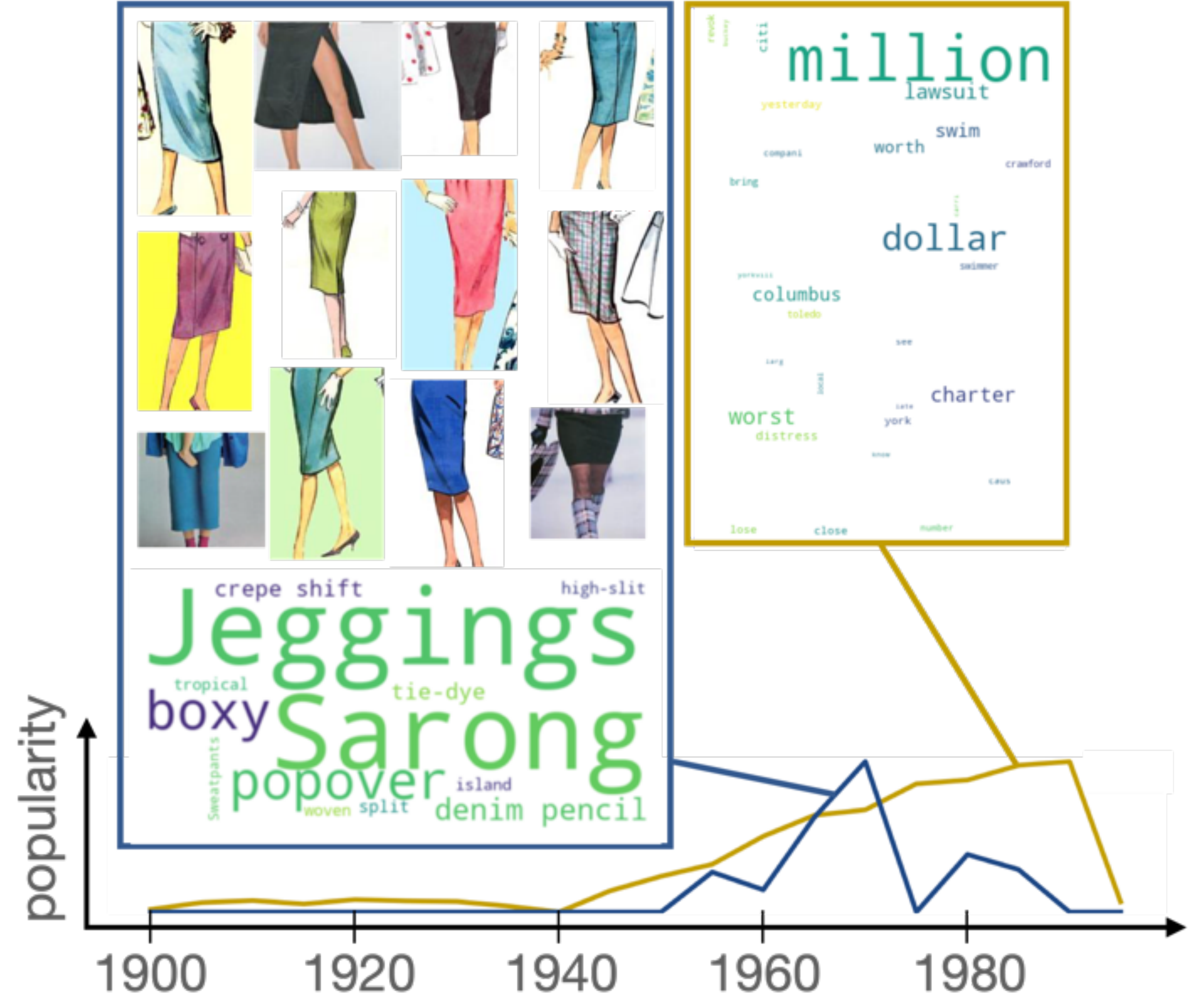}
  }\subfloat[Topic '\emph{Conference}' influences stylish-business clothing.\label{fig:conference_influence}]{
    \includegraphics[width=.31\linewidth]{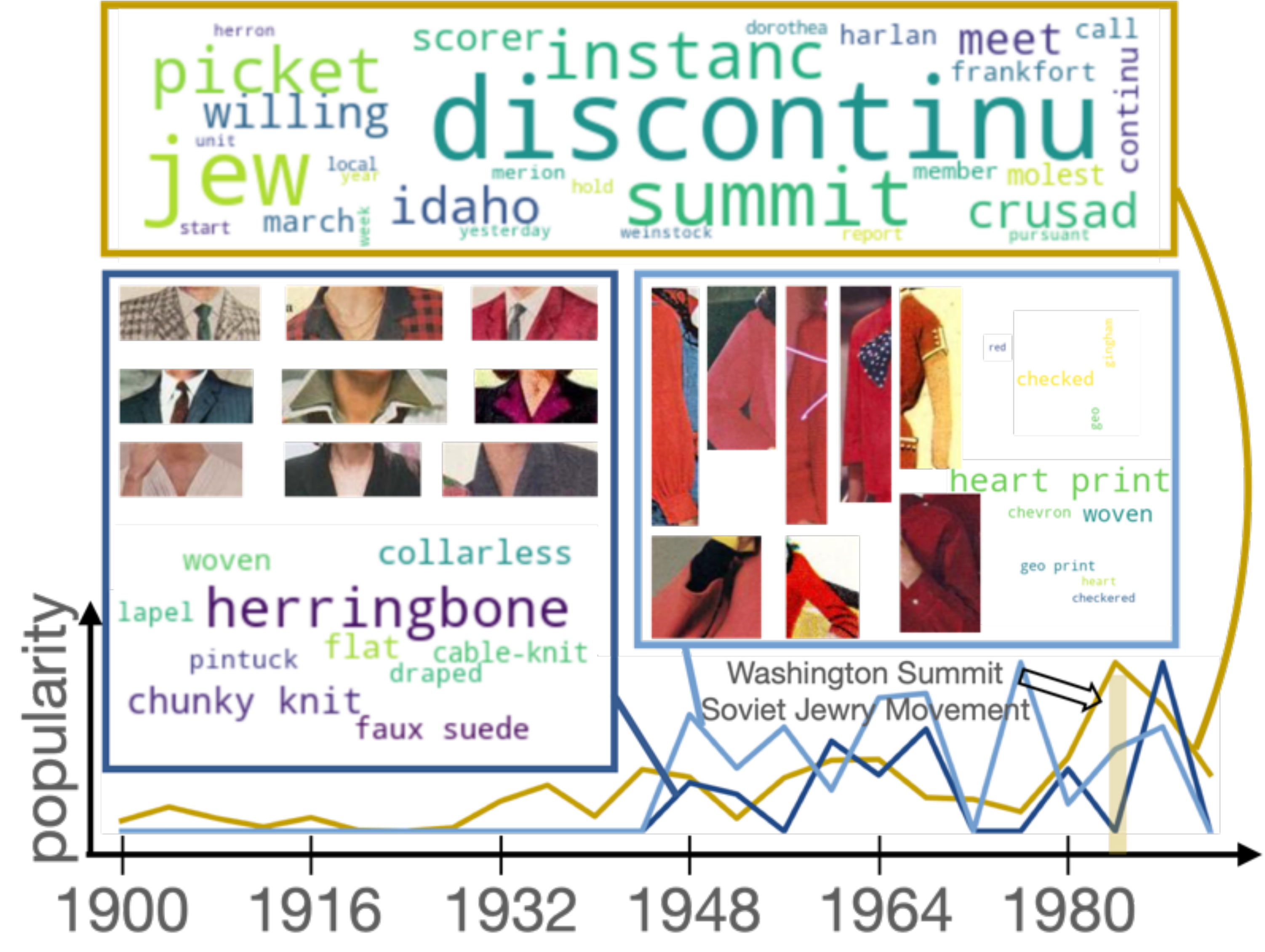}
  }\subfloat[Topic `\emph{Love}` influences bridal styles.\label{fig:love_influence}]{
    \includegraphics[width=.37\linewidth]{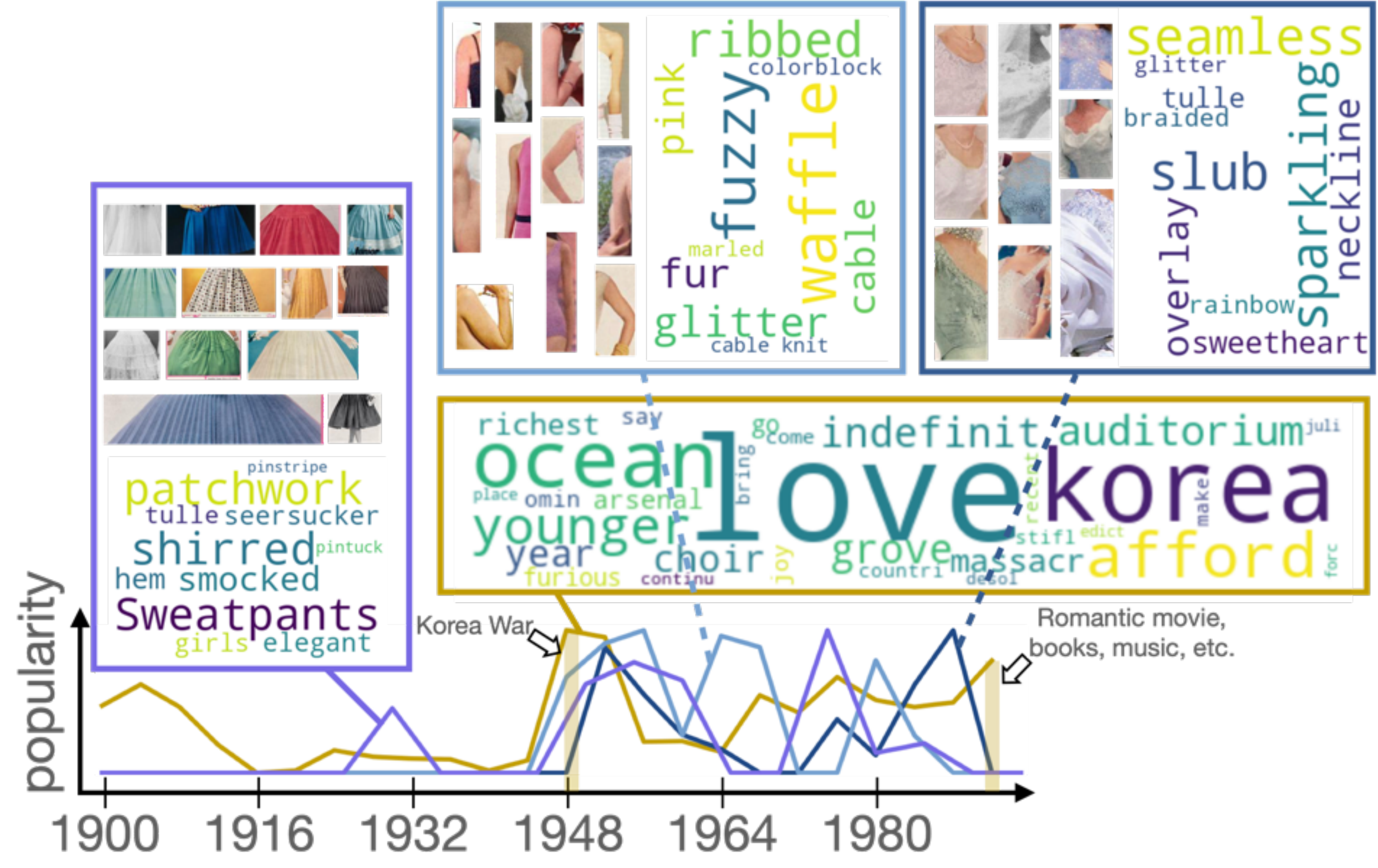}
  }\\
  \subfloat[`\emph{Fashion}` influences dresses in Paris.\label{fig:fashion_influence}]{
    \includegraphics[width=.33\linewidth]{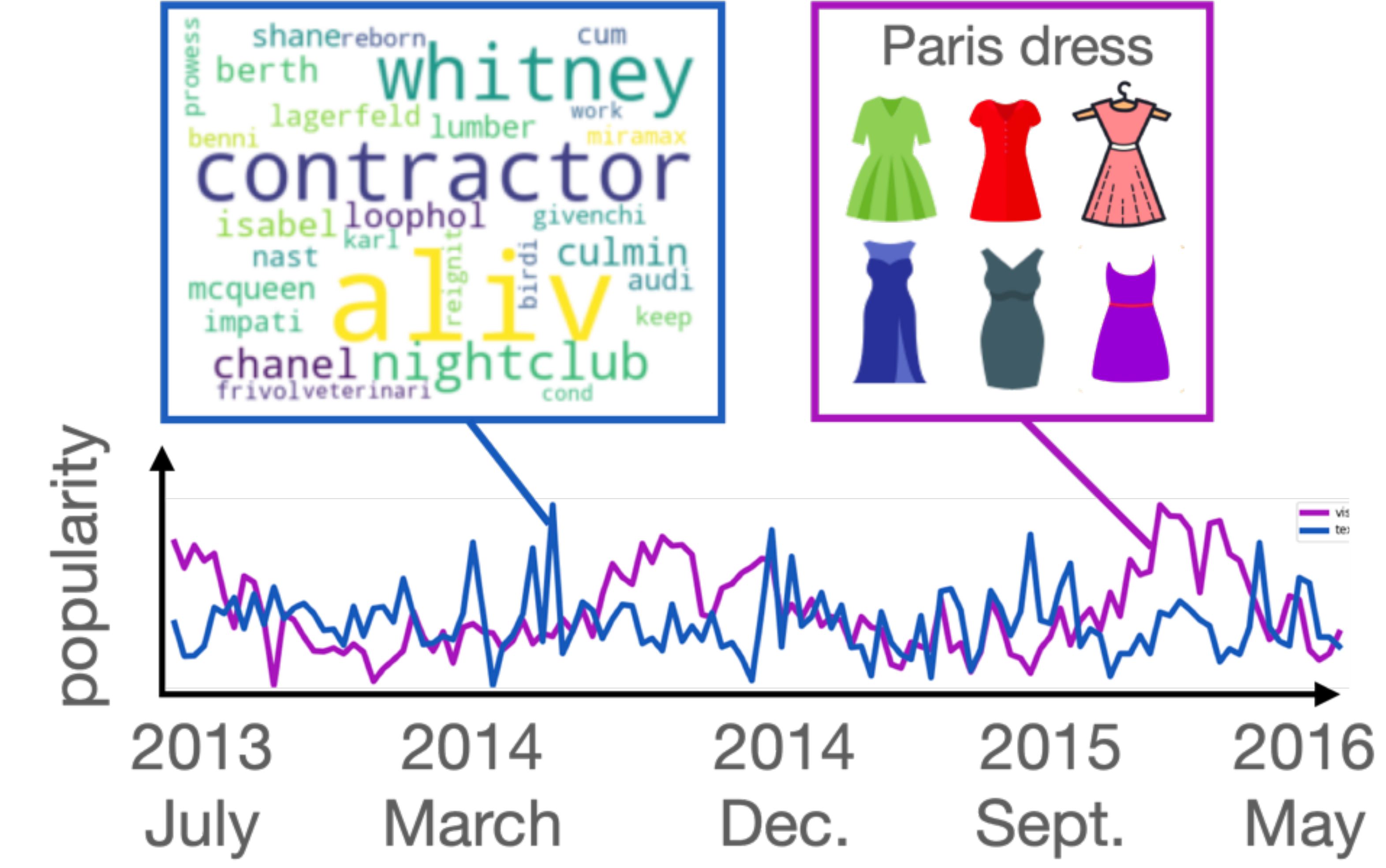}
  }\subfloat[`\emph{Music}` influences wearing scarves in Milan.\label{fig:music_influence}]{
    \includegraphics[width=.33\linewidth]{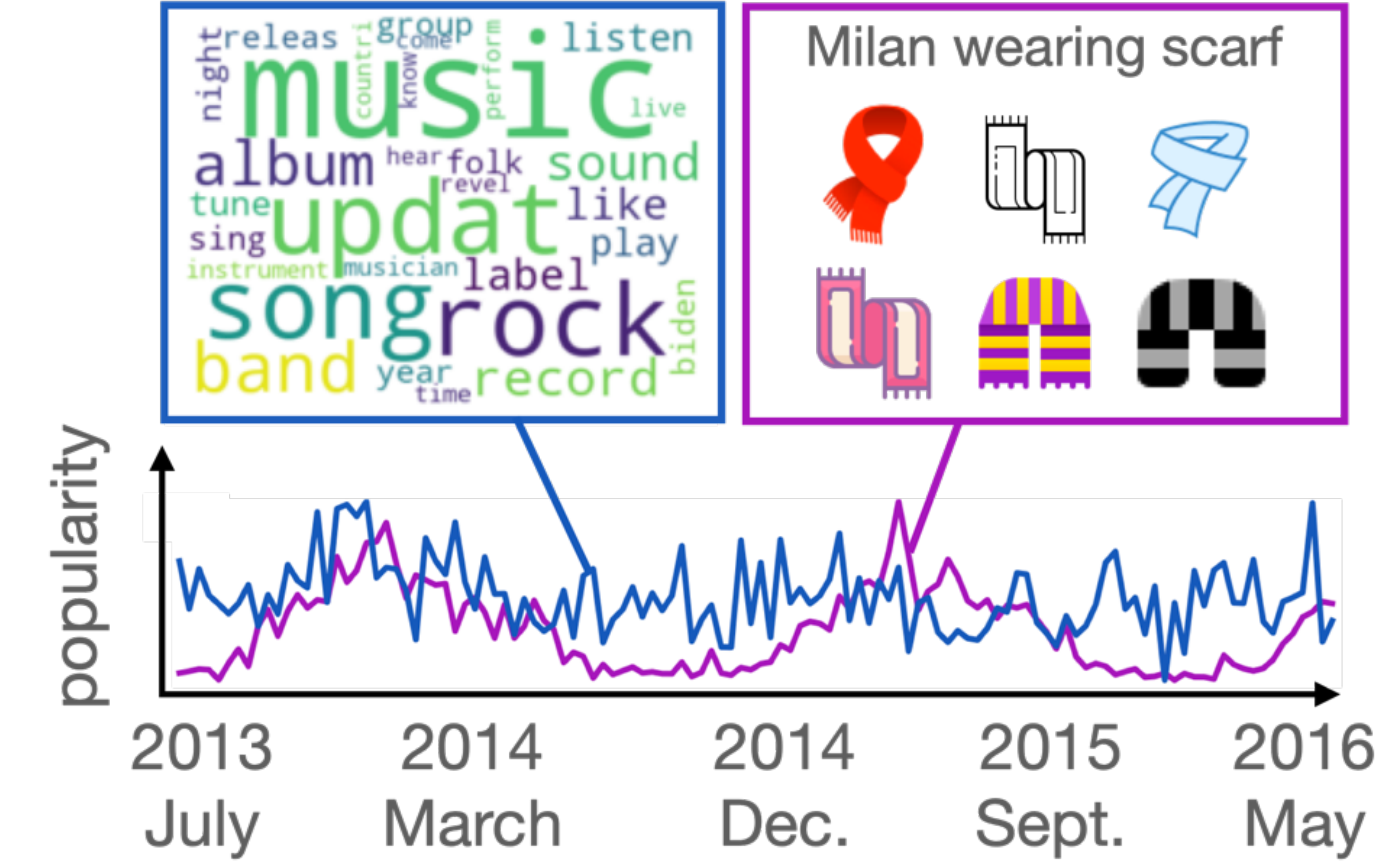}
  }\subfloat[`\emph{Sports}` influences blue clothing in Madrid.\label{fig:sports_influence}]{
    \includegraphics[width=.33\linewidth]{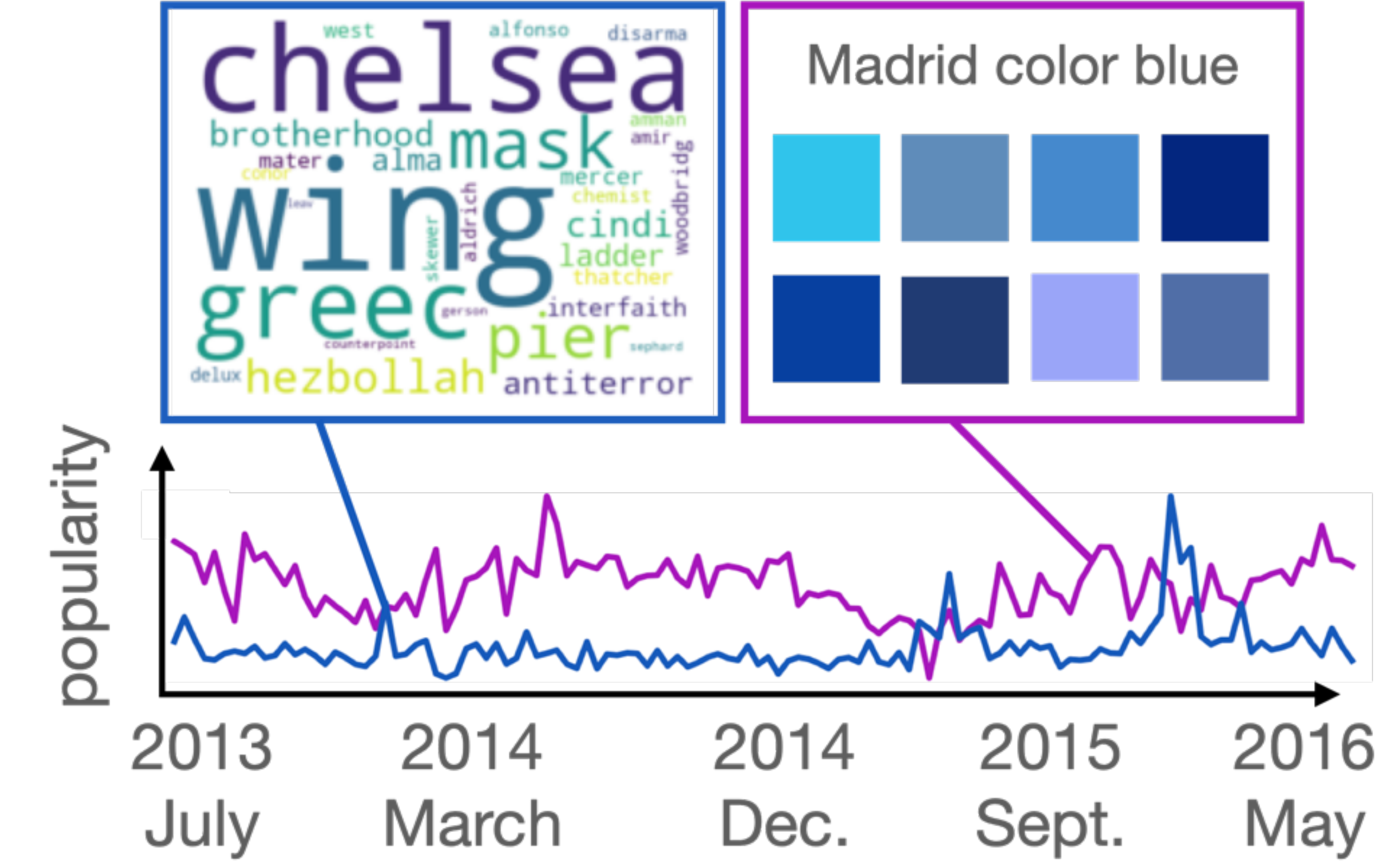}
  }
  \caption{\textbf{More examples of discovered influences}. Fig.\ (a-c) are on the Vintage dataset, while Fig.\ (d-f) are on the GeoStyle dataset.}
  \label{fig:more_discovered_influences} 
\end{figure*}

\begin{figure*}
  \subfloat[]{
    \includegraphics[width=.45\linewidth]{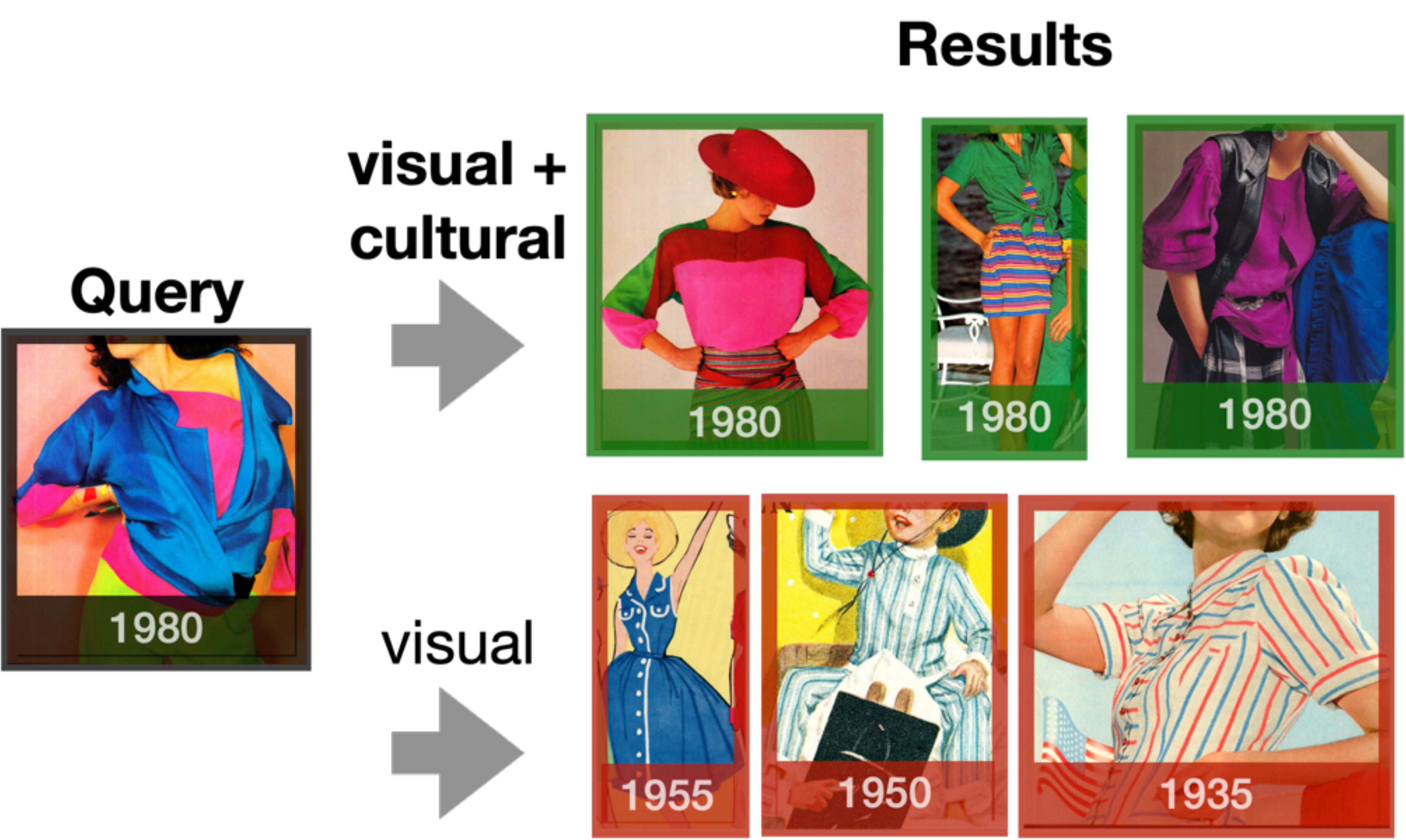}
  }\hfill\subfloat[]{
    \includegraphics[width=.45\linewidth]{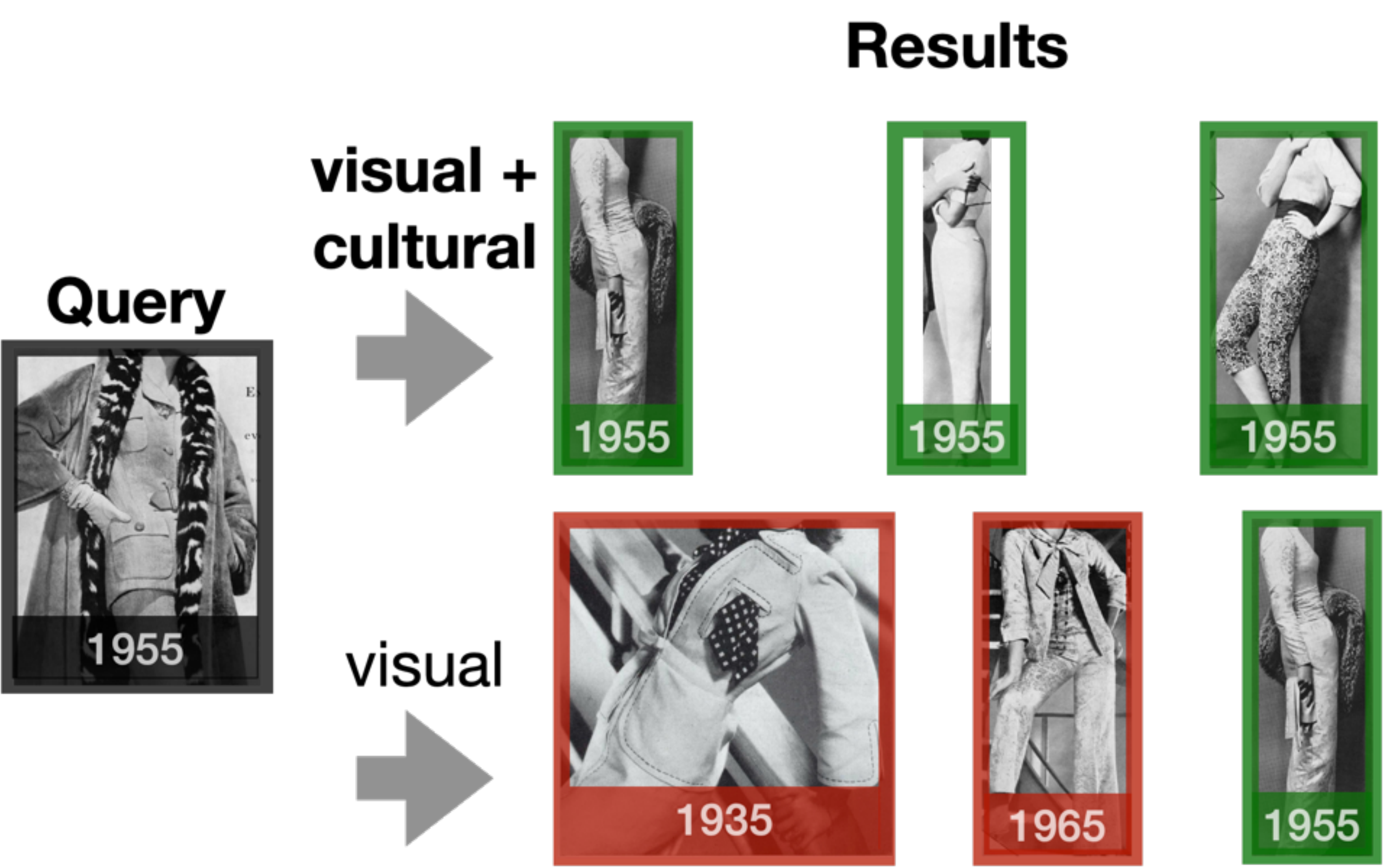}
  }\caption{\textbf{Examples of timestamping through retrieval}: Query instance is on the left, and retrieved results are on the right. The top row in each example is retrieved using visual features augmented with our inferred cultural features, while the bottom row uses visual features only. 
  True date labels are shown on the bottom of each photo. 
  Temporally consistent photos retrieved are bounded by green boxes, 
  while temporally inconsistent ones are bounded by red boxes. 
  While all retrieved results are stylistically similar (color, pattern, fit, \etc) to the query, 
  those retrieved by visual with cultural features capture more temporally-sensitive styles, \ie, styles unique at that time.}
  \label{fig:timestamp_ex} 
\end{figure*}

\section{Examples of failure cases in influence-based forecasting}
\begin{center}
    \begin{figure*}[t]
      \begin{minipage}[c]{0.12\linewidth} 
        \includegraphics[width=\linewidth]{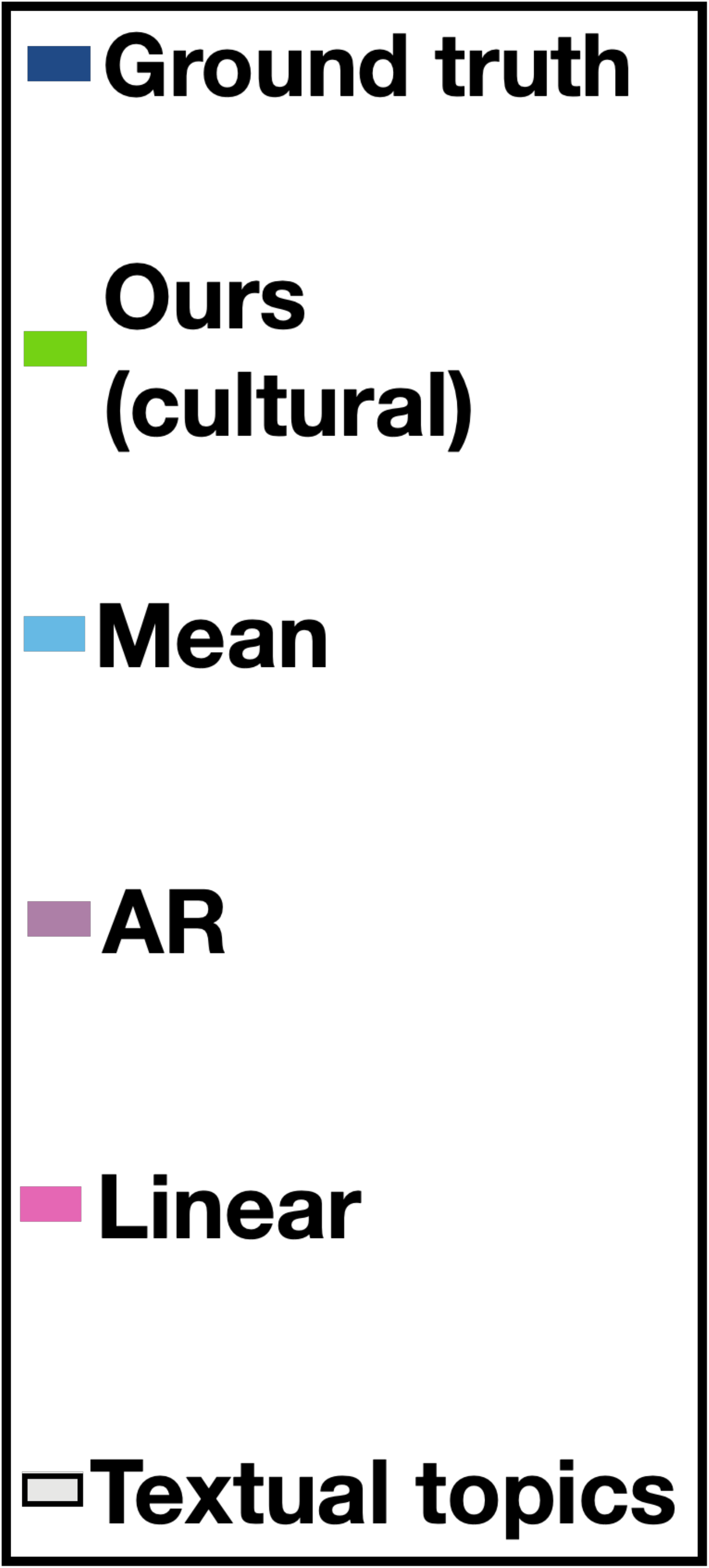}
      \end{minipage}
      \hspace{1mm}
      \begin{minipage}{0.83\linewidth}
        \raisebox{6mm}{(a)} 
        \includegraphics[width=.48\linewidth]{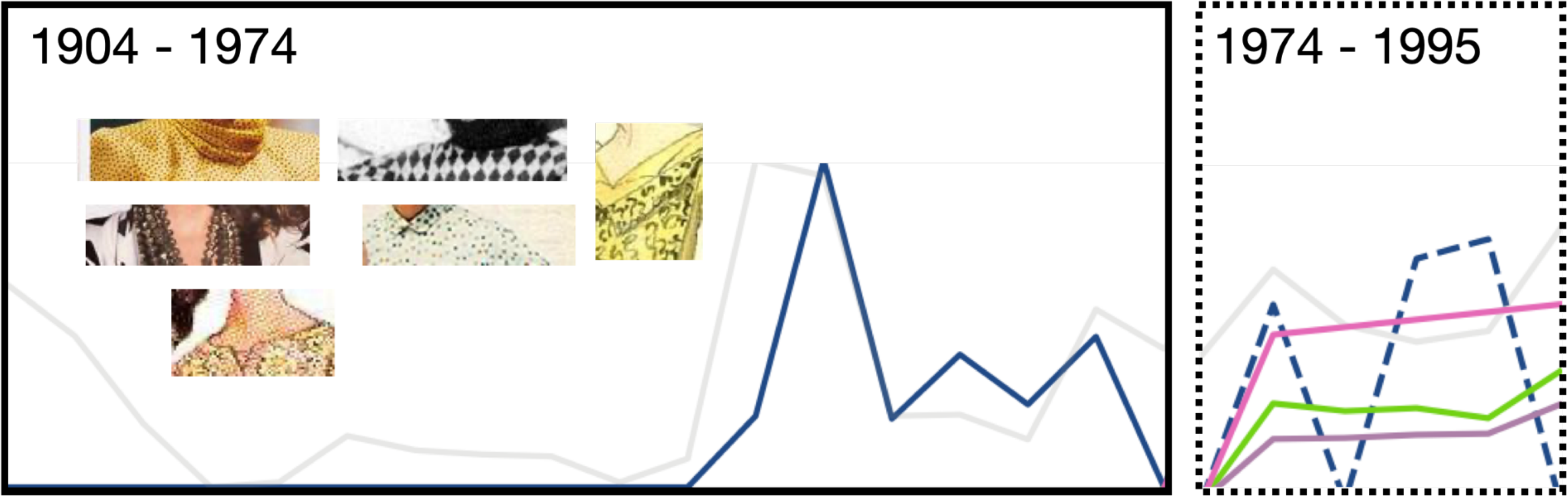}
        \raisebox{6mm}{(b)}  
        \includegraphics[width=.48\linewidth]{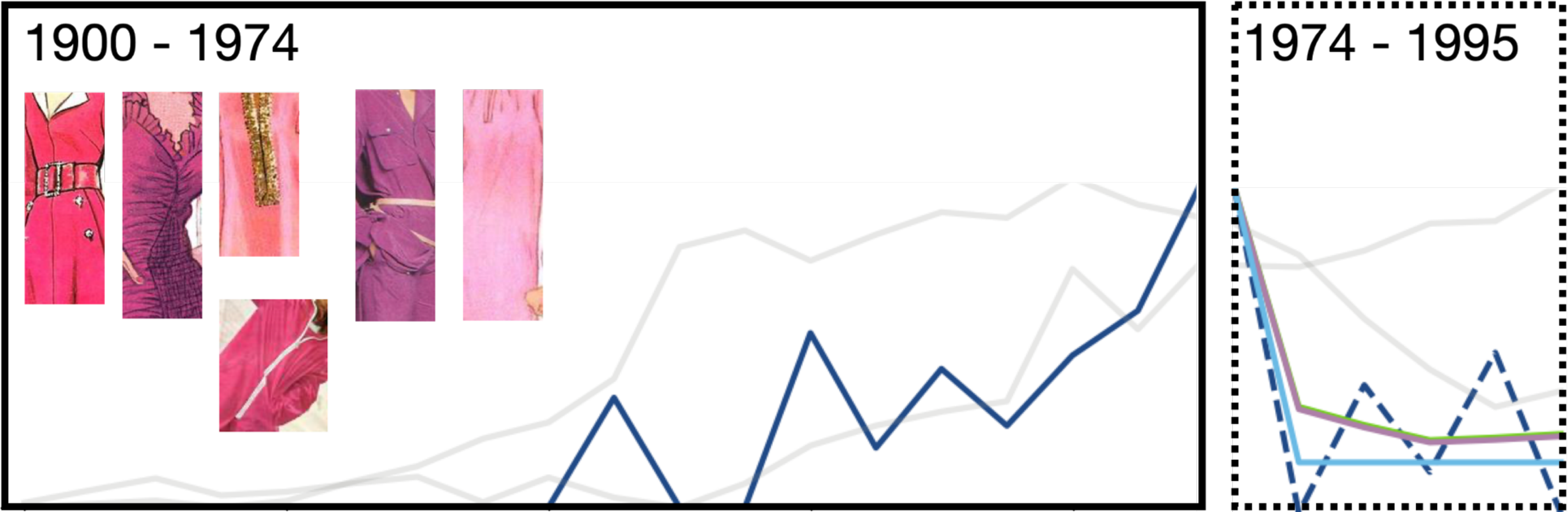}
        \\
        \raisebox{6mm}{(c)}  
        \includegraphics[width=.48\linewidth]{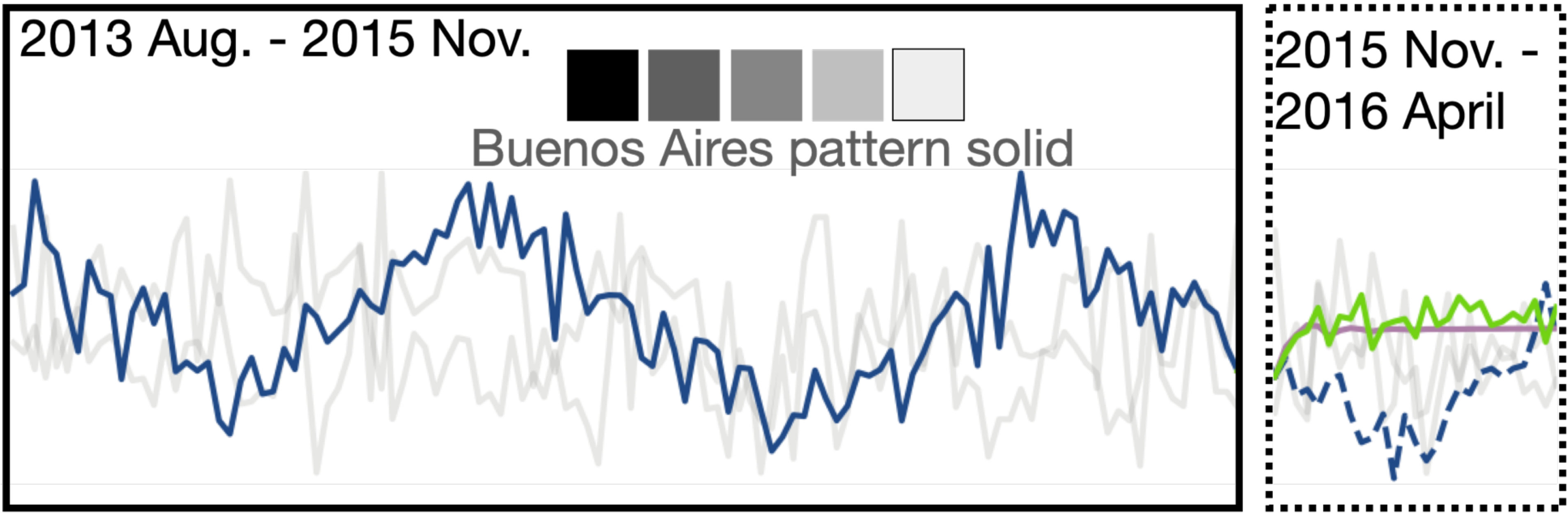}
        \raisebox{6mm}{(d)}  
        \includegraphics[width=.48\linewidth]{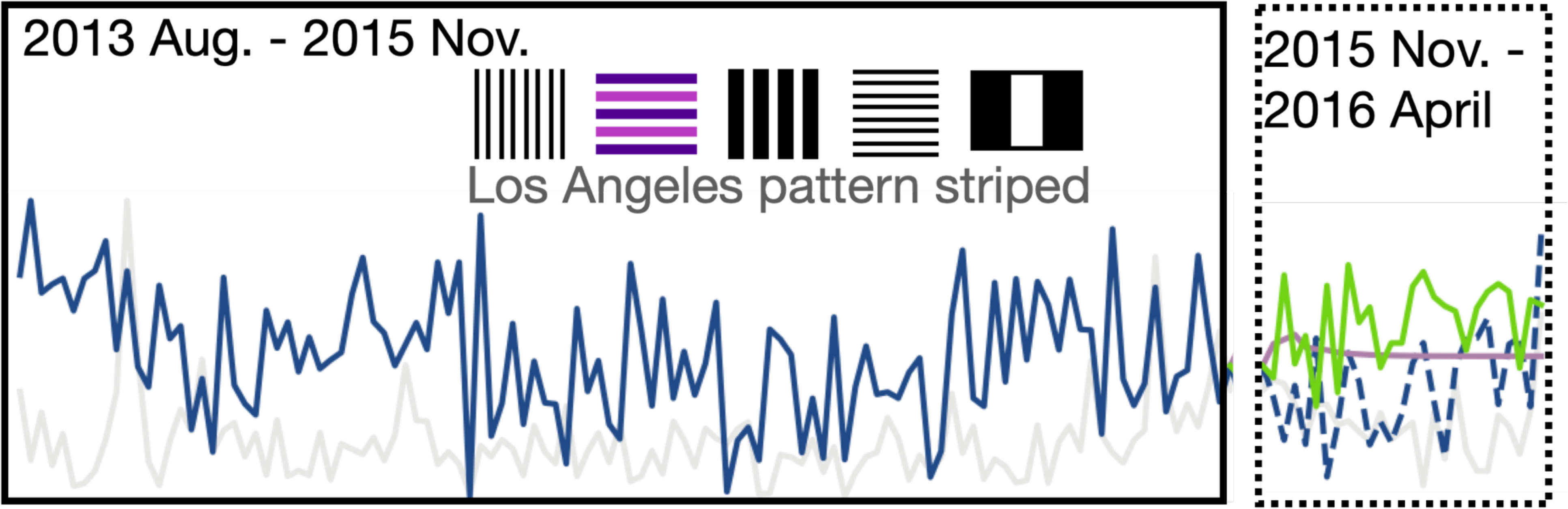}
      \end{minipage}\hspace{-10mm} \\
      \begin{minipage}[c]{0.98\linewidth}  
        \vspace{6mm}  
        \caption{\textbf{Examples of failure cases} when considering cultural influences in trend forecasting. 
                Top row on the Vintage data, and bottom row on the GeoStyle data. 
                Future trends predicted by including cultural influences in AR are inaccurate, 
                because the cultural time series did not offer useful information during test time. 
                In some of these cases, more naive baselines like mean or linear predictions perform better than both vanilla AR or AR including cultural infelunces.}
        \label{fig:trend_predict_failure} 
      \end{minipage}
      \vspace*{-1mm}
    \end{figure*}
\end{center}

\KH{Quantitative results for applying our detected influences on trend-forecasting are in Tab.1.\ in the main paper, 
and qualitative examples of how cultural influences help predict more accurate trends are in Fig.6.\ in the main paper. 
In summary, including cultural influences in autoregression (AR) to predict future trends improves $57\%$ of the styles on the Vintage data when comparing to AR, 
and improves $80\%$ of the styles on the GeoStyle data. 
\figref{trend_predict_failure} shows qualitative examples when including cultural influences could not help AR (\figref{trend_predict_failure}(a-b)), 
or perform even worse (\figref{trend_predict_failure}(c-d)). 
Trend forecasting is an extremely challenging task, 
so cultural influences detected on the training series may no longer hold true on the test series, 
sometimes containing no useful information, 
or even outliers that result in unstable predictions. 
In those cases, more naive baselines like mean (\figref{trend_predict_failure}(b)) or linear (\figref{trend_predict_failure}(a)) predictions perform better than both vanilla AR and AR including cultural influences. %
}

\section{User study for visual cluster's quality}
\begin{figure*}
    \centering
    \includegraphics[width=.99\linewidth]{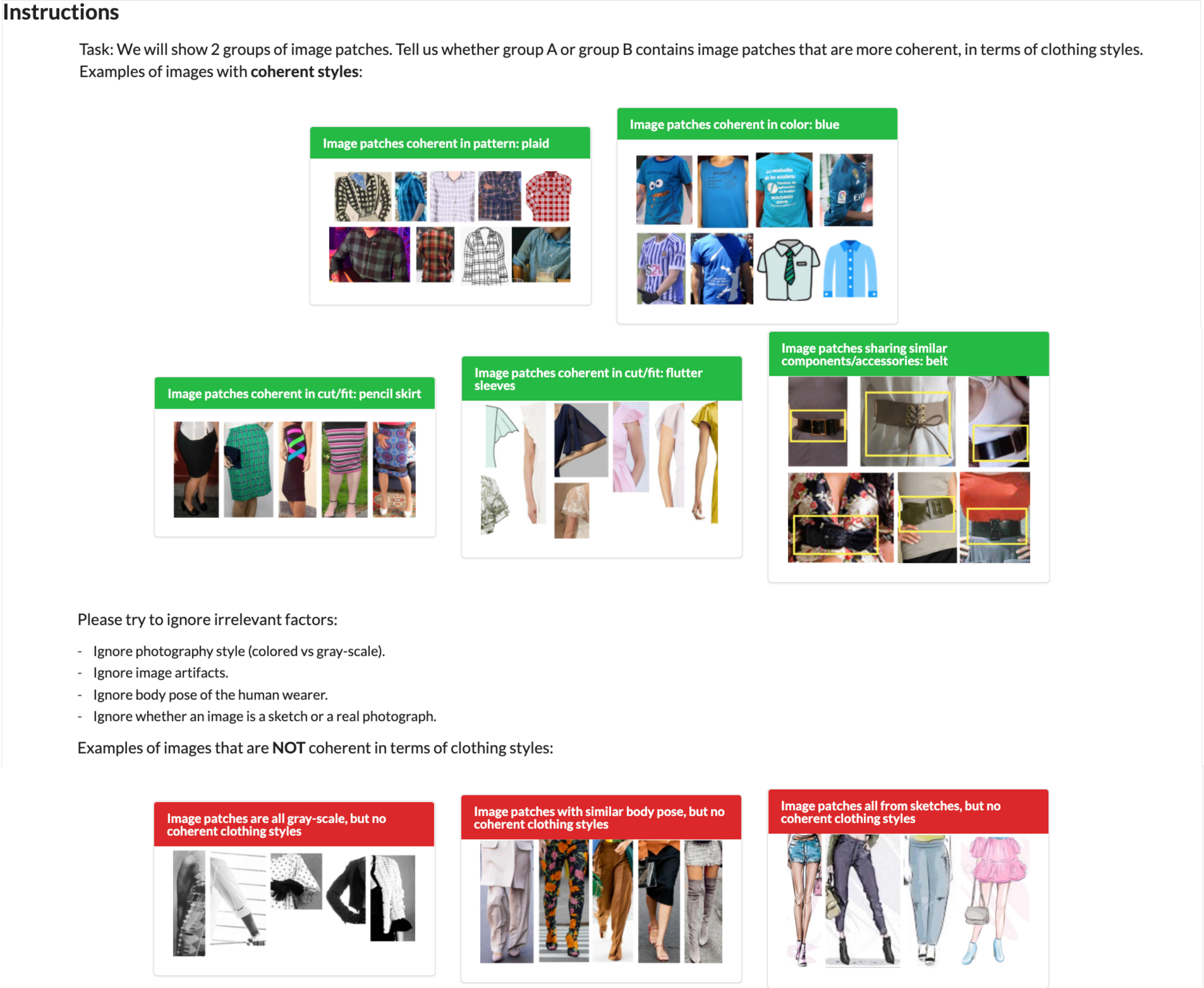}
    \caption{Instruction page for user study on whether a visual cluster consists of coherent clothing styles.}
    \label{fig:amt_instruction} 
\end{figure*}

\begin{figure*}
    \centering
    \includegraphics[width=.99\linewidth]{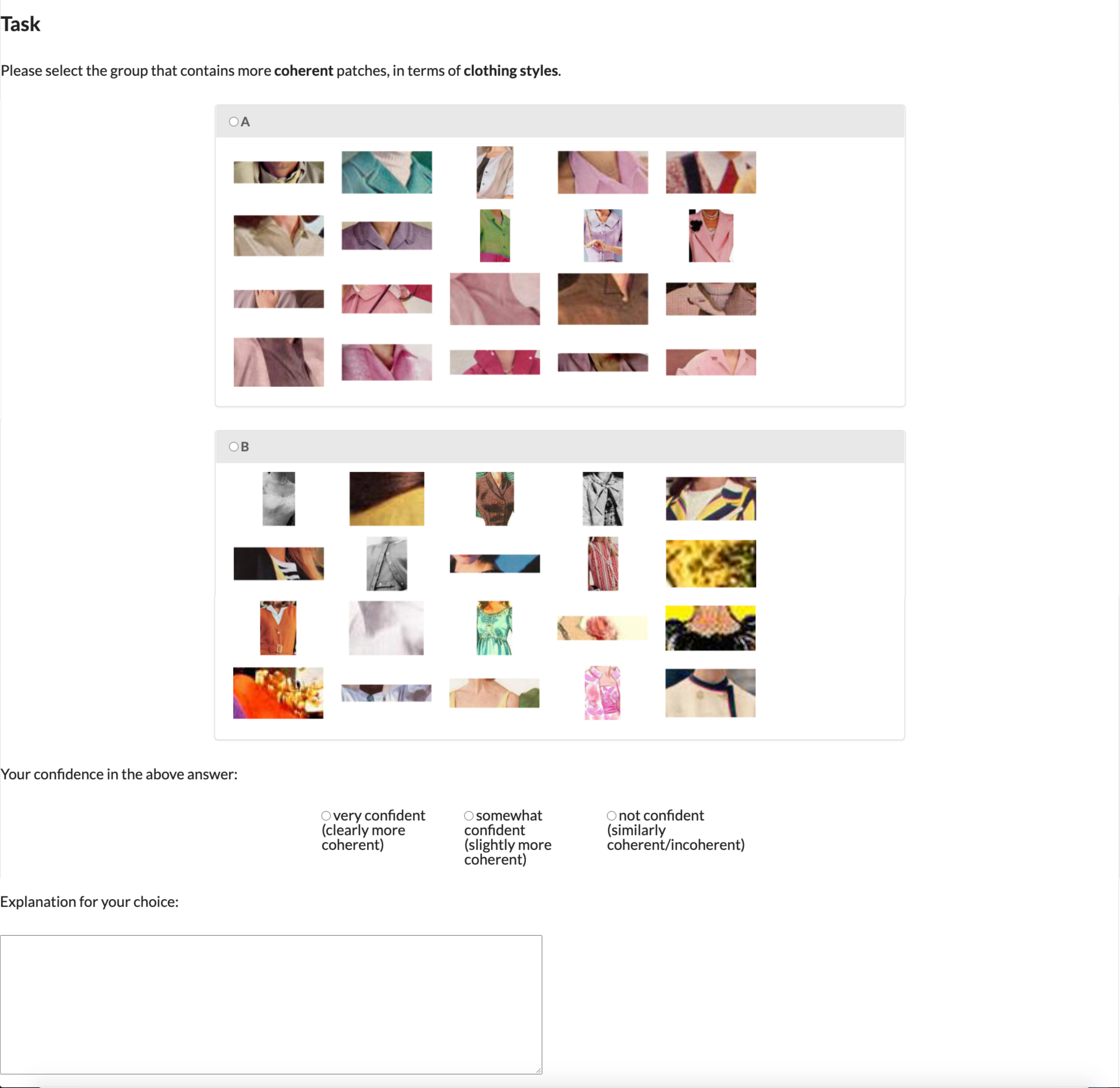}
    \caption{Task page for user study on whether a visual cluster consists of coherent clothing styles: one of the options is our algorithm's discovered cluster, the other is a random grouping of images.}
    \label{fig:amt_question} 
\end{figure*}

To verify whether the automatically discovered visual clusters correspond to meaningful clothing styles, we conduct a user study with the following protocol:
for each visual cluster, we compare its most centroid $20$ images with $20$ randomly sampled images (as shown in \figref{amt_question}), 
and ask human subjects to select the option that is more coherent in terms of \emph{clothing styles}. 
Our instruction clarifies which factors are relevant or irrelevant to clothing styles (as shown in \figref{amt_instruction}). 

The user study is conducted on the Amazon Mechanical Turk platform, 
and each pair of comparison is answered by $5$ to $7$ Turkers, 
in total $156$ unique Turkers. 

$75\%$ of the time, human judges find the clusters to exhibit coherent clothing styles that they can describe (as reported in the main paper), 
and example descriptions they gave for what they see in the selected images are:
`\emph{Most of the pictures include sleeveless tops for women. Most of the models in the picture seem to wear a necklace.}', 
`\emph{Group A is more coherent because many patches in it have formal suits.}', 
`\emph{A group is more coherent because of the pink and pastel color prevalence.}', 
`\emph{I chose B because some patches show jackets or dresses with long sleeves.}', 
`\emph{Image patches are very coherent in terms of showing sleeves of garments, mostly short sleeves. Most patches of clothing appear to be cut from a fine fabric such as linen or silk and expensive-looking.}'

The human judges not only find that most of our automatically discovered clusters are coherent, 
but explanations for their decisions also show that they were not based on differences in photography techniques. 
The fact that human judges can see and describe the coherence of the styles discovered by our method is evidence that we do find meaningful styles despite the very wide span of time (100 years) in the Vintage photos.

\end{document}